\documentclass[sigconf,review=false]{acmart}

\usepackage{multirow}
\usepackage{makecell}
\usepackage{lineno,hyperref}

\AtBeginDocument{%
  }

\begin{document}

\title{OpenEP: Open-Ended Future Event Prediction}

\author{Yong Guan}
\affiliation{%
  \institution{Tsinghua University}
  \state{Beijing}
  \country{China}
  }
\email{gy2022@mail.tsinghua.edu.cn}

\author{Hao Peng}
\affiliation{%
  \institution{Tsinghua University}
  \state{Beijing}
  \country{China}
  }
\email{peng-21@mail.tsinghua.edu.cn}

\author{Xiaozhi Wang}
\affiliation{%
  \institution{Tsinghua University}
  \state{Beijing}
  \country{China}
  }
\email{wangxz20@mail.tsinghua.edu.cn}

\author{Lei Hou}
\affiliation{%
  \institution{Tsinghua University}
  \state{Beijing}
  \country{China}
  }
\email{houlei@tsinghua.edu.cn}

\author{Juanzi Li}
\affiliation{%
  \institution{Tsinghua University}
  \state{Beijing}
  \country{China}
  }
\email{lijuanzi@tsinghua.edu.cn}

\renewcommand{\shortauthors}{Yong Guan et al.}

\begin{abstract}
 
Future event prediction (FEP) is a long-standing and crucial task in the world, as understanding the evolution of events enables early risk identification, informed decision-making, and strategic planning. Existing work typically treats event prediction as classification tasks and confines the outcomes of future events to a fixed scope, such as yes/no questions, candidate set, and taxonomy, which is difficult to include all possible outcomes of future events. In this paper, we introduce \textbf{OpenEP} (an \underline{Open}-Ended Future \underline{E}vent \underline{P}rediction task), which generates flexible and diverse predictions aligned with real-world scenarios. This is mainly reflected in two aspects: firstly, the predictive questions are diverse, covering different stages of event development and perspectives; secondly, the outcomes are flexible, without constraints on scope or format. To facilitate the study of this task, we construct \textbf{OpenEPBench}, an open-ended future event prediction dataset. For question construction, we pose questions from seven perspectives, including time, location, event development, event outcome, event impact, event response, and other, to facilitate an in-depth analysis and understanding of the comprehensive evolution of events. For outcome construction, we collect free-form text containing the outcomes as ground truth to provide semantically complete and detail-enriched outcomes. Furthermore, we propose \textbf{StkFEP}, a stakeholder-enhanced future event prediction framework, that incorporates event characteristics for open-ended settings. Our method extracts stakeholders involved in events to extend questions to gather diverse information. We also collect historically events that are relevant and similar to the question to reveal potential evolutionary patterns. Experiment results indicate that accurately predicting future events in open-ended settings is challenging for existing LLMs. In addition, we thoroughly summarize the problems encountered in prediction, hoping to provide insights for future research.

\end{abstract}



\maketitle

\section{Introduction}

\begin{figure}
    \centering
    \includegraphics[width=\linewidth]{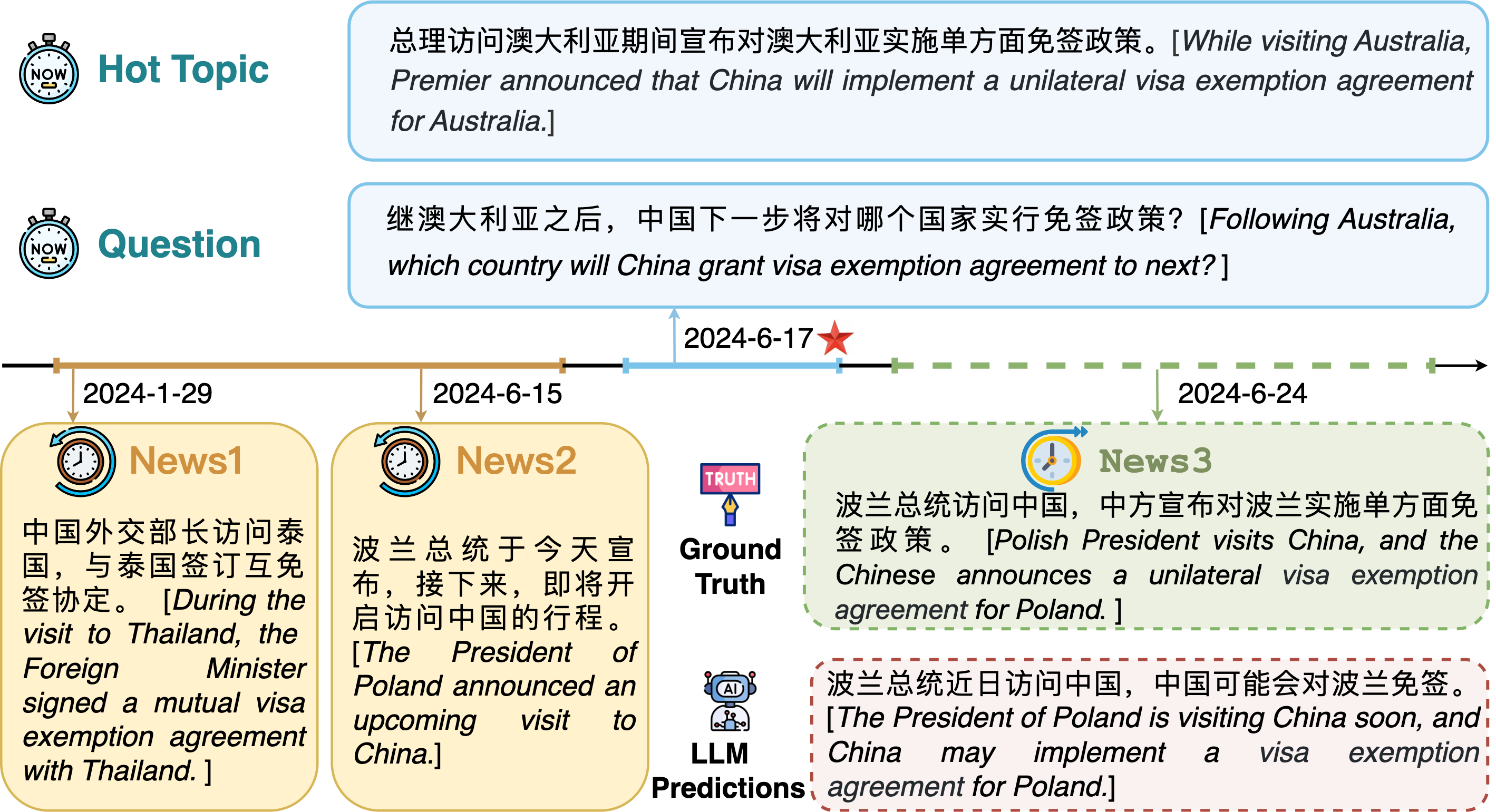}
    \caption{Example of Future Event Prediction.}
    \label{fig:intro_sample}
\end{figure}

Future event prediction (FEP) aims to predict the potential future outcomes based on the historical events and the precursors~\cite{zhao2021event}. 
In Figure \ref{fig:intro_sample}, given the predictive question ``\textit{Which country will China grant visa exemption agreement to next?}'', a FEP model needs to collect historical events, such as ``\textit{The President of Poland announced an upcoming visit to China}'', to predict the outcomes of future events like ``\textit{Poland}''. 
Accurate anticipation of future events is crucial in the modern world, as it provides scientific support for making more rational and efficient decisions and possesses significant practical and application value. Such as, in fields law~\cite{pumi_pred_sigspatial_2018}, finance~\cite{yang_financial_gnn_cikm_2019}, and healthcare~\cite{pmlr-v70-dempsey17a}, event prediction technology helps identify potential risks and uncertainties, enhancing safety and risk management capabilities through logical analysis and predicting.

Early research mainly utilizes statistical machine learning methods, fitting predefined statistical models with historical data for prediction~\cite{laxman_stream_kdd_2008,hashimoto-etal-2014-toward}. These methods often required the integration of domain knowledge 
%
and involved complex feature engineering. With the rapid advancement of big data and deep learning technologies, predicting through data-driven neural networks has emerged as an appealing alternative~\cite{li-etal-2021-future,NEURIPS2022_aec870a6,ma_forcast_kdd_2023}. Especially in recent years, the emergence of large language models (LLMs)~\cite{openai2023gpt4,glm2024chatglmfamilylargelanguage} have exhibited astonishing performance in tasks previously thought to require human cognitive abilities. 
Despite some work focusing on LLM-based event prediction~\cite{schoenegger2024wisdomsiliconcrowdllm,pratt2024languagemodelsuseforecasting,halawi2024approachinghumanlevelforecastinglanguage,ye2024miraievaluatingllmagents}, \textbf{open-ended future event prediction task is still being neglected}. Existing work typically treats event prediction as classification tasks and confines the outcomes of future events to a fixed scope, such as yes/no questions~\cite{schoenegger2024wisdomsiliconcrowdllm,halawi2024approachinghumanlevelforecastinglanguage,ye2024miraievaluatingllmagents}, candidate set~\cite{zhu_generative_aaai_2023,bai_rich_aaai_2023,ma_context_kdd_2023}, and taxonomy~\cite{li-etal-2021-future}. The fixed scope of outcomes typically comprises a uniform format, such as single words or short phrases, resulting in predicted outcomes that often lack rich semantics and details. In contrast, freely generated outcomes in real-world usually contain longer and semantically complete responses, enriched with more details.

In this paper, we introduce \textbf{OpenEP} (an \underline{Open}-Ended Future \underline{E}vent \underline{P}rediction task), which generates more flexible and diverse predictions aligned with real-world scenarios. This is mainly reflected in two aspects. (1) The predictive questions are diverse, covering different stages of event development and perspectives, facilitating a comprehensive analysis. (2) The outcomes are flexible, without constraints on scope, format, or length, which can provide semantically complete responses enriched with more details. 
To facilitate the study of this task, we first construct \textbf{OpenEPBench}, an open-ended future event prediction dataset, and concurrently propose \textbf{StkFEP}, a stakeholder-enhanced future event prediction framework for open-ended settings.

For OpenEPBench construction, we need to address the following three key questions: \textit{(Q1)} How to determine the data source? \textit{(Q2)} How to generate predictive questions? \textit{(Q3)} How to annotate the outcomes of future events? 
\textit{For data source, } we select hot topics from two platforms widely used for discussing daily events, utilizing Zhihu for Chinese data and Google News for English data.  
\textit{For question generation, } we pose questions from seven perspectives, including location, time, event development, event outcome, event impact, event response, and other, to facilitate in-depth analysis and understanding of the comprehensive evolution of events. 
\textit{For outcome annotation, } due to the advanced comprehension capabilities of LLMs, they can evaluate event predictions from a semantic perspective just like human evaluators. Therefore, we extract segments from the original texts containing outcomes as ground truth, without employing a fixed scope or format. In addition, we design corresponding LLM-based evaluation metrics, which measure the predictions from five dimensions, including accuracy, completeness, relevance, specificity, and reasonableness.

For framework StkFEP, it contains three modules: \textit{Retrieval}, \textit{Integration}, and \textit{Prediction}. 
The \textit{Retrieval} aims to collect the diverse information from news sources to mitigate the semantic gap between the question and predictions. 
The evolution of events depends on salient entities involved, regard as \textit{stakeholders}. Knowing these entities aids in question expansion and facilitates the retrieval of diverse information. For instance, in Figure \ref{fig:intro_sample}, given the stakeholders \textit{China} and \textit{premier} can effectively retrieve the news ``\textit{The President of Poland announced an upcoming visit to China}''. Thus, we extract stakeholders to extend original questions to gather the diverse information. 
In addition to retrieving news directly related to the question, referred to as \textit{relevant events}, we also collect historically occurred events that are similar to the question, known as \textit{similar events}. These similar events can serve as references for predicting future events. For example, by considering news 1, it can be inferred from news 2 whether China will sign a visa exemption agreement with Poland. 
The \textit{Integration} module employs clustering method to clarify the dependencies between events and reduce redundant information. 
At last, the \textit{Prediction} aims to predict the outcomes based on the information of relevant and similar events. 
During the testing phase, tests are conducted immediately after daily question annotations are completed to minimize the risk of information leakage. 
To summarize our main contributions:
\begin{itemize}
    \item We introduce \textbf{OpenEP}, an open-ended future event prediction task that generates flexible and diverse predictions aligned with real-world scenarios.
    \item We construct \textbf{OpenEPBench}, an open-ended future event prediction dataset with diverse predictive questions and flexible outcomes, facilitating comprehensive analysis. In addition, we design LLM-based metrics to evaluate the model predictions.
    \item We propose \textbf{StkFEP}, a stakeholder-enhanced future event prediction framework that incorporates the characteristics of event evolution for open-ended settings. 
    \item Extensive experiments demonstrate that accurately predicting future events in open-ended settings is challenging for existing LLMs. Furthermore, we have thoroughly summarized the problems encountered in prediction.

\end{itemize}

\section{OpenEPBench}

In this section, we will describe the OpenEPBench dataset. First, we introduce the overall dataset construction (Sec~\ref{sec_overview}). Next, we introduce the construction process, which includes the data source (Sec.~\ref{sec_data_source}), constructing the predictive questions (Sec.~\ref{sec_question_cons}), and their corresponding outcomes (Sec.~\ref{sec_outcome_cons}). 
Once the dataset is built, we analyze its distribution and perform quality checks (Sec.~\ref{sec_data_analysis}). Finally, we introduce the evaluation metrics (Sec.~\ref{sec_evaluation}). 
Further construction and annotation details, including annotation interfaces, examples, and annotation guidelines, are provided in Appendix~\ref{appendx_data_construction}.

\subsection{Overview}
\label{sec_overview}

This section aims to introduce the overall dataset construction procedure. Our goal is to build an open-ended FEP dataset featuring diverse predictive questions and flexible outcomes that are unconstrained by scope or format. Predictive questions are annotated on a daily basis, and outcomes are collected at future time points. The model is tested daily after the predictive questions are constructed, and it is evaluated when the outcomes are collected.

Prediction window is the time span from the current moment into which future events or values are projected. 
Considering that people's attention to hot topics generally lasts around 7 days~\cite{kwak2010twitter} and that attention to major health emergencies can extend to over 13 days~\cite{liu2022event}. Therefore, we set the prediction window to 15 days, predicting events that might occur within the next 15 days.

Data annotation, while labor-intensive and costly due to the substantial resources and domain expertise required, ensures high accuracy. The advent of advanced LLMs, exemplified by GPT-4~\cite{openai2023gpt4}, offers a transformative opportunity for the data annotation process. Consequently, we employs a combination of LLMs and human verification to construct the dataset. The LLMs automate initial annotations, significantly reducing manual labor, while human checks ensure the accuracy and relevance of the annotations. The data construction process consists of two stages: Question Construction and Outcome Construction. 
For \textit{Question Construction}, collect daily hot topics, use the LLMs to generate multiple potential predictive questions for each hot topic, and manually validate and filter these questions. 
For \textit{Outcome Construction}, after the prediction window for individual question, use LLMs to collect news within the period, score the news articles, and then manually validate the event outcomes.

\begin{figure}
    \centering
    \includegraphics[width=\linewidth]{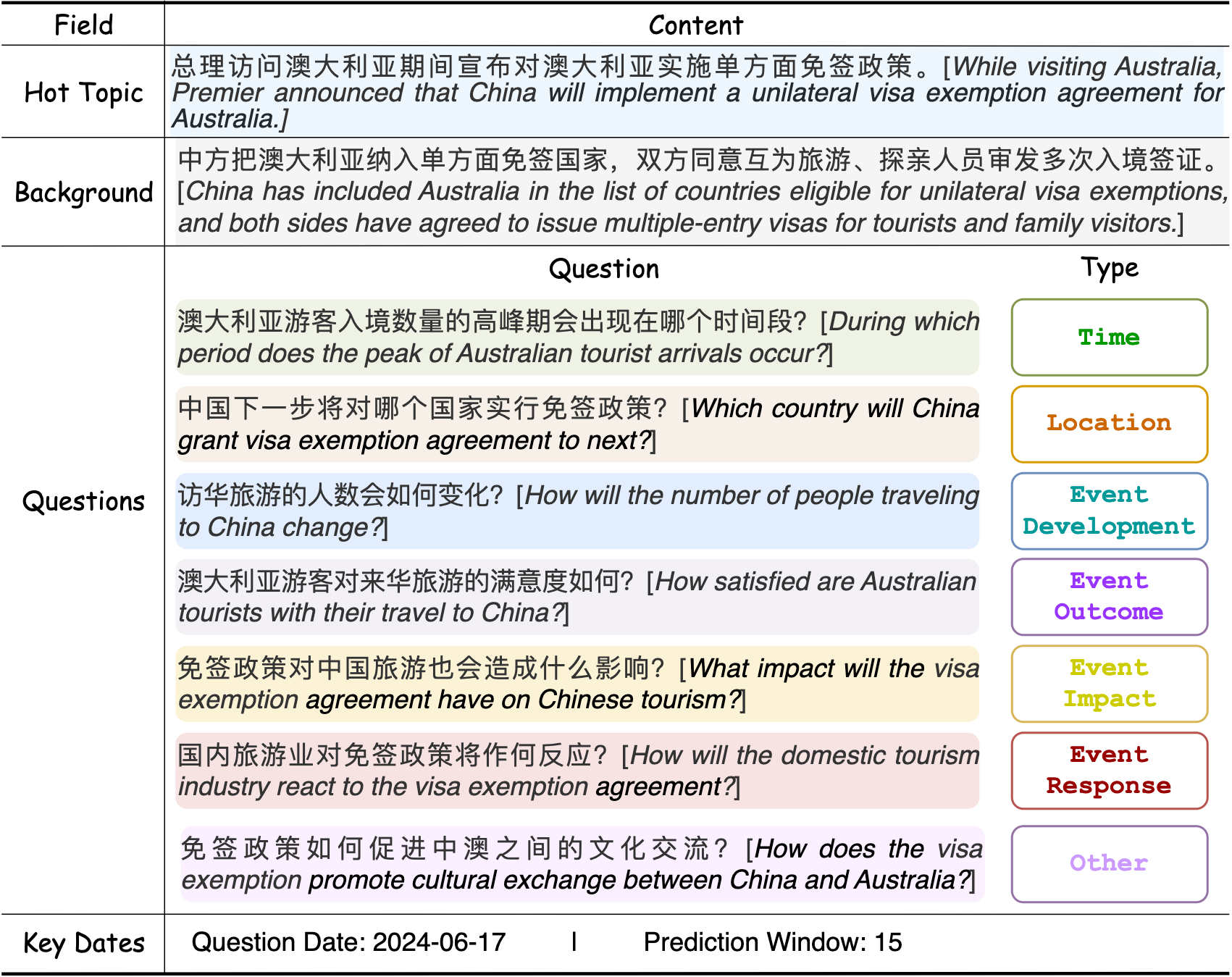}
    \caption{Example from the OpenEPBench dataset.}
    \label{fig:display_example}
\end{figure}

\subsection{Data Source}
\label{sec_data_source}

Data Source aims to identify and select reliable and relevant sources of data. 
To construct a prediction dataset that aligns with real-world scenarios, we utilize news data from the internet, which is continuously updated and comprehensive. 
LLMs are trained with vast amounts of data across various languages, giving them multilingual understanding capabilities. However, their performance still varies across different languages. Therefore, to test the robustness of these models, we have constructed separate datasets in Chinese and English. 
We focus on constructing predictive questions based on daily hot topics. We have chosen two widely used platforms for this purpose: Zhihu~\footnote{\url{https://www.zhihu.com/knowledge-plan/hot-question/hot}} for Chinese data, where daily discussions on hot topics are directly utilized as hot topics, and Google News~\footnote{\url{https://news.google.com/topics}} for English data, where the headlines of news are regard as hot topics.

\subsection{Question Construction}
\label{sec_question_cons}

Question Construction aims to build predictive questions based on hot topics. 
The overall process involves collecting hot topics, using LLMs to generate candidate questions, and then manually verifying the candidate questions. 

Events evolve dynamically and undergo various phases. By formulating predictive questions from multiple perspectives, we can conduct a more thorough analysis and understanding of the event evolution. Therefore, incorporating the elements of an event 5W1H (who, where, when, what, why, how)~\cite{sharma2013news}, along with external feedback, we propose posing questions from seven perspectives about hot topics, as shown in Figure \ref{fig:display_example}.

\begin{itemize}
    \item Time. The date on which a future event is likely to occur.
    \item Location. The specific place or location where a future event will occur.
    \item Event Development. The progression of a future event, including how the event unfolds, potential movements, or turning points.
    \item Event Outcome. The direct results or outcomes after a future event has concluded.
    \item Event Impact. The impact of a future event on the surrounding environment, economy, or other relevant sectors.
    \item Event Response. The reactions of different stakeholders to an event, including the public, governments, markets, or specific groups' behavioral and emotional responses.
    \item Other. Any additional aspects or perspectives that require further clarification.
\end{itemize}

\textbf{Construction process.} Question construction employs a combination of LLMs and manual verification, comprising the following five steps: 
(1) \textit{Hot Topics Collection}. Collect daily hot topics from Zhihu and Google News.
(2) \textit{Background Collection}. Hot topics collected online include a background typically generated from a single news article. To enrich the background, we retrieve related news articles using the hot topic and use LLMs to regenerate a background for supplementation.
(3) \textit{Hot Topic Validity Verification}. Not all hot topics are suitable for generating questions. For instance, many hot topics may involve discussions of past events that have resurfaced. Hence, based on LLMs, we verify each hot topic for aspects such as continuity, initially filtering out those that do not meet the criteria.
(4) \textit{Candidate Question Generation}. For each hot topic, the LLM generates multiple predictive questions from the previously mentioned seven perspectives. Each perspective may contain multiple questions.
(5) \textit{Human Verification}. Manually verify questions generated by the LLM based on answerability, specificity, and real-time, selecting suitable predictive questions.

\begin{figure}
    \centering
    \includegraphics[width=0.6\linewidth]{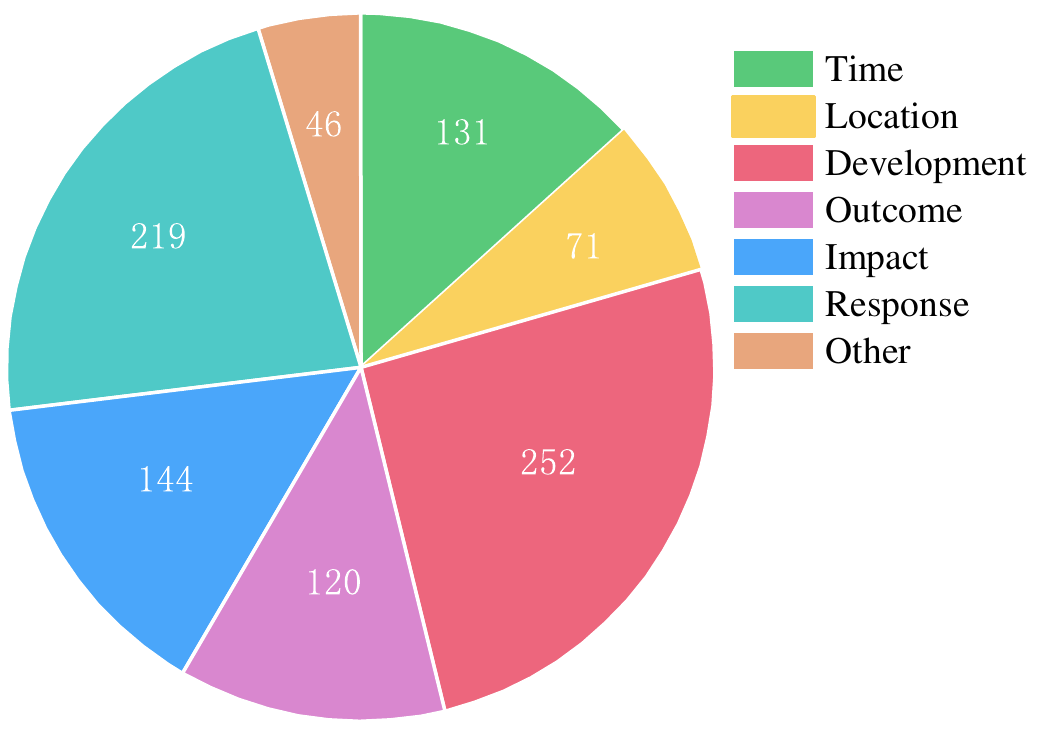}
    \caption{Data distribution across different types of questions.}
    \label{fig:question_distribution}
\end{figure}

\subsection{Outcome Construction}
\label{sec_outcome_cons}

Outcome Construction aims to collect the ground truth for the corresponding predictive questions. The overall process involves using LLMs to collect news, score the articles, and manually verify the outcomes. 
The outcomes of future events are not constrained by fixed scopes, formats, or lengths. Relying solely on manual generation of event outcomes undoubtedly increases the workload. Interestingly, LLMs have inherent strengths in comprehension and generation, effectively grasping contextual semantic information. Currently, LLMs are widely used to evaluate model performance~\cite{narsupalli2024reviewfeedbackreasonrefernovelframework,zhang2024largelanguagemodelsevaluators}, providing a semantic perspective that surpasses traditional assessments like ROUGE or BLEU, which measure word co-occurrence. Therefore, we extract segments from news articles that contain the outcomes as ground truth.

\textbf{Construction process.} Outcome construction utilizes a combination of LLMs and manual verification, consisting of the following three steps: 
(1) \textit{News Collection}. For each question, retrieving news from news sources within the prediction window. 
(2) \textit{News Rerank}. Each news article is scored by LLMs to assess if it contains the event outcomes, with higher scores indicating a greater probability. The news articles are then ranked by the scores, and the valid news are selected as candidates. 
(3) \textit{Human Verification}. The selected news articles are manually verified to extract segments containing the event outcomes, forming the final ground truth.

\subsection{Data Analysis}
\label{sec_data_analysis}

\noindent \textbf{Data quality. } 
To ensure data quality, for automatic annotation, we select GPT-4, currently the best-performing LLM. For human annotation, we invited two experts with PhDs in natural language processing to help check the data. The specific process involves each individual independently verifying the results from the LLM during both the question construction and outcome construction phases. After validation, any inconsistencies in the data are discussed between the two experts. Data agreed upon through discussion is accepted, while data with unresolved discrepancies is discarded.

\noindent \textbf{Data Distribution. } 
%
For Chinese data, from June 1, 2024, to July 10, 2024, we collect 192 valid hot topics over a 40-day period and generate 869 valid predictive questions, with an average of 4.52 predictive questions per hot topic.  For English data, from July 1, 2024, to July 10, 2024, we collect 27 valid hot topics over a 10-day period and generate 114 valid predictive questions, with an average of 4.22 predictive questions per hot topic.

The questions in our dataset cover a very wide variety of perspectives. We design seven types of predictive questions, including time, location, event development, event outcome, event impact, event response, and other. Figure \ref{fig:question_distribution} shows the data distribution across each predictive question category.

\subsection{Evaluation Metrics}
\label{sec_evaluation}

The predictive questions include seven types: time, location, event development, event outcome, event impact, event response, and other. When evaluating time-type questions, we convert the question into a multiple-choice format, using accuracy as the evaluation criterion. Specifically, the prediction window is divided into three periods, each covering five days, plus an additional option indicating no outcome, creating a total of four options. The prediction model must output one of these options as the result. 

Apart from time-type prediction questions, the outcomes for other types of questions are presented in free-form text, without constraints on scope, format, or length. Traditional automatic evaluation metrics, which measure word co-occurrence, are no longer suitable, necessitating a more human-like approach to assessment from a semantic perspective. More recently, LLMs exhibit astonishing performance in tasks previously thought to require human cognitive abilities and are increasingly used to evaluate model performance. Inspired by existing work~\cite{narsupalli2024reviewfeedbackreasonrefernovelframework,zhang2024largelanguagemodelsevaluators}, we utilize LLMs, such as GPT-4, to evaluate event prediction performance from the following five dimensions: 
\begin{itemize}
    \item Accuracy. Measures the extent to which the predicted content matches the actual outcomes or states that occurred, with a primary focus on the precision of the predictions. 
    \item Completeness. Assesses whether the prediction covers the different relevant aspects of the actual outcomes, evaluating the thoroughness of the information provided.
    \item Relevance. Evaluates how pertinent the prediction is to the actual outcomes, ensuring that the prediction does not veer into unrelated details.
    \item Specificity. Analyzes the sharpness and focus of the prediction, ensuring that it is neither overly broad nor vague. 
    \item Reasonableness. Measures the logical coherence and believability of the prediction, checking whether the prediction aligns with general world knowledge and appears plausible.
\end{itemize}

When measuring \textit{Accuracy}, for each prediction question, the actual outcomes may contain multiple aspects of information. If the prediction hits at least one aspect, it scores 1; otherwise, it scores 0. For \textit{Completeness}, the score is calculated as the proportion of accurately predictions relative to the actual outcomes. For the other three dimensions—\textit{Relevance}, \textit{Specificity}, and \textit{Reasonableness}—scores for each dimension are on a scale from 1 to 5, where higher scores indicate better performance. Existing research indicates that LLMs can be overconfident~\cite{xiong2024can,yang2024trustllmsmitigateoverconfidence}. Therefore, when LLMs provide scores, we require them to also offer probabilities. The final score for each dimension is calculated as follows:
\begin{equation}
    \mathit{score}= \sum_{i=1}^{n} \sigma(s_{i}) * \rho(s_i)
\end{equation}
where $\sigma(\cdot)$ aims to map the score to the range 0-1, $\rho(\cdot)$ represents the probability of the score, and $n$ denotes the size of the data. 

Ultimately, we aggregate the scores from different dimensions and question types to gauge overall performance. In addition to automatic evaluation, we also conduct human evaluations on a subset of the entire dataset. Similar to automatic evaluation, human evaluators provide scores from the aforementioned dimensions.

\begin{figure*}
    \centering
    \includegraphics[width=0.9\linewidth]{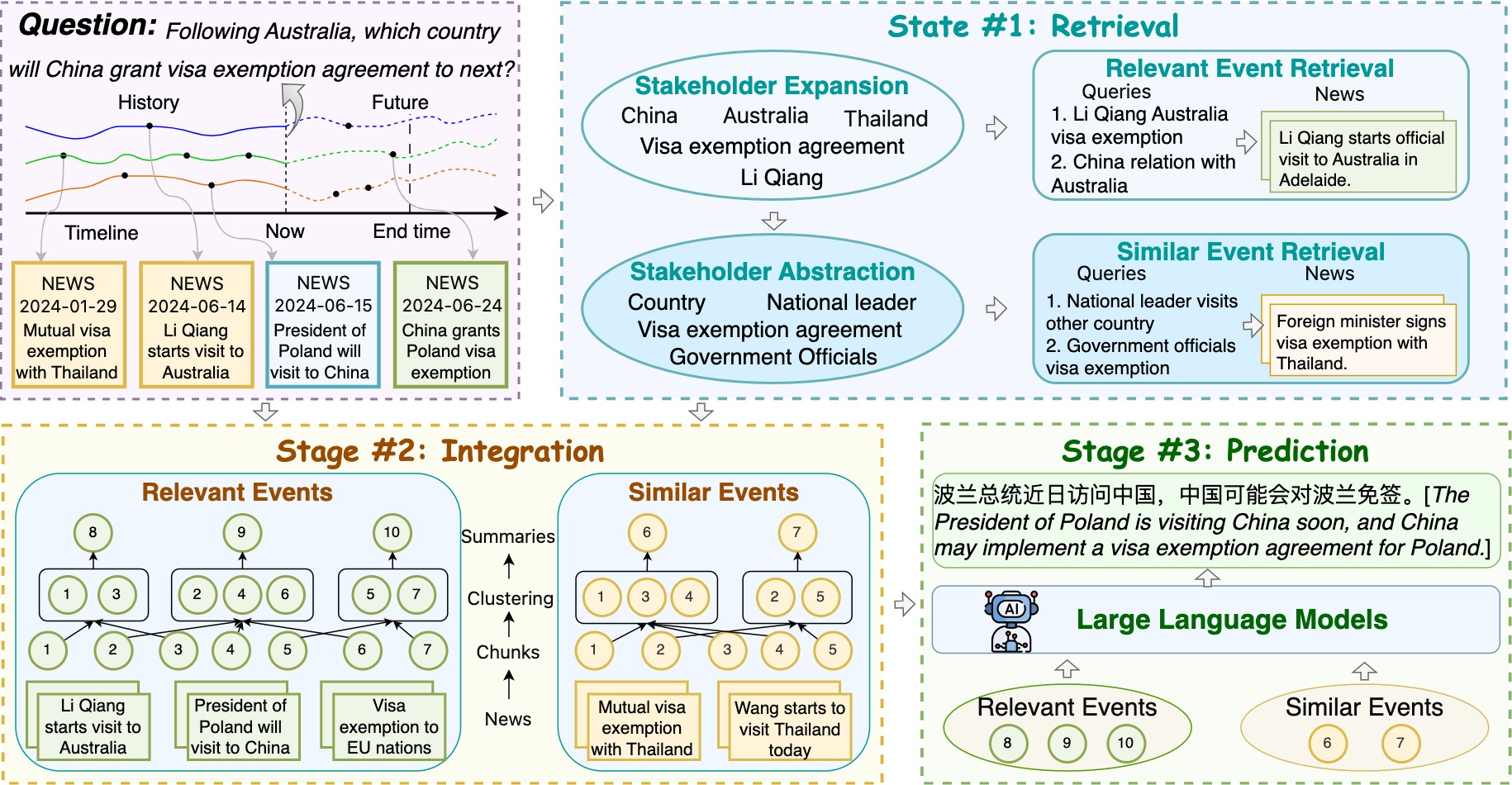}
    \caption{The framework of StkFEP.}
    \label{fig:fep_framework}
\end{figure*}

\section{StkFEP}

In this section, we introduce StkFEP, a stakeholder-enhanced future event prediction framework for open-ended settings. We first present the task description (Sec.~\ref{sec_task_desc}). Next, we detail the StkFEP framework, comprising three modules: Retrieval (Sec.~\ref{sec_retrieval}), Integration (Sec.~\ref{sec_integration}), and Prediction (Sec.~\ref{sec_prediction}), as depicted in Figure~\ref{fig:fep_framework}.

\subsection{Task Description}
\label{sec_task_desc}

Each item $x_{i}$ in the dataset $\mathcal{D}$ can be represented as a quintuple $x_{i} = (q, t, w, b, o)$, where $q$ is the predictive question, $t$ is the time the question is built, noted as a timestamp\footnote{Each timestamp represents a day, formatted in ``YYYY-MM-DD''.}, $w$ is the prediction window, $b$ is the background of the question, and $o$ is the actual outcomes of the question. 
Relevant events $SE$ refer to news information that is directly related to the question. Similar events $RE$ are historically occurred events that are similar to the question. Assume $q^{\prime}$ is the expanded set of question $q$, $f$ is the prediction system, and $o^{\prime}$ represents the predicted outcomes of $f$.

Given $q$ and $b$, the prediction system $f$ is required to expand $q$ into $q^{\prime}$  at time $t$, using both $q$ and $q^{\prime}$ to retrieve $SE$ and $RE$ from news sources. Based on $SE$ and $RE$, the system predicts potential outcomes $o^{\prime}$. The performance of the model is then evaluated after constructing $o$ following the prediction window $w$.

\subsection{Retrieval}
\label{sec_retrieval}

The Retrieval module aims to collect diverse information from news sources to support the prediction. It consists of 3 steps: question expansion, relevant event retrieval, and similar event retrieval. 

\noindent \textbf{Question Expansion. } This module aims to expand the original question to facilitate the retrieval of diverse information. 
The information retrieved using the original question is insufficient for event prediction, necessitating the expansion of the question. However, existing methods mainly focus on the capabilities of LLMs, allowing these models to autonomously generate multiple questions while overlooking the characteristics of event. 
The evolution of events depends on salient entities involved, regard as \textit{stakeholders}~\cite{kuila_stak_sigir_2024}. Knowing these entities aids in question expansion and enhances information retrieval. Therefore, we extract stakeholders to extend the original questions and gather comprehensive information. 
Specifically, we first use the original question to retrieve news from news sources and prompt the LLM to assess the relevancy and filter out irrelevant news. Then, extracting stakeholders from each news article. Based on the original question, background, and stakeholders, we use the LLM to generate various questions. 

\noindent \textbf{Relevant Event Retrieval. } 
This module aims to retrieve relevant events based on the expanded questions. Relevant events are those directly related to the predictive question and can help provide a comprehensive background for the question. News are retrieved from news sources based on both the expanded questions and original question. However, not all retrieved news articles are relevant. To filter out irrelevant news, we utilize the LLM to score each article and remove those with low scores.

\begin{table*}[t]
    \centering
        \caption{Model performance of different types of questions on Chinese data (\%). Detailed experimental results for each dimension can be found in the appendix \ref{appendix_indi_dim}.}
    \begin{tabular}{ll|cccccccc}
    \toprule
    Models&Methods&Time&Location&Development&Outcome&Impact&Response&Other&Overall\\
    \midrule 
    \multirow{4}*{GPT-3.5}&DR + Summ&32.65&47.83& 41.22& 45.39& 46.18& 37.65& 38.86& 37.52\\
    &DR + Summ-o-Summ&38.77&46.21& 43.06& 46.81& 47.68& 37.37& \textbf{44.43}& 41.12\\
    &GQR + Summ-o-Summ&39.79&48.73& 43.49& 47.29& 46.28& 38.51& 41.03& 43.78\\
    &StkFEP&\textbf{45.92}&\textbf{49.21}& \textbf{45.88}& \textbf{50.84}&\textbf{ 54.26}& \textbf{38.93}& 40.53& \textbf{46.95}\\
    \midrule
    \multirow{4}*{GLM-4}&DR + Summ&38.65&28.33& 34.92& 40.70& 39.07& 32.90& 35.24& 37.08\\
    &DR + Summ-o-Summ&40.18&35.68& 35.69& 42.57& 40.92& 34.44& 34.02& 38.72\\
    &GQR + Summ-o-Summ&42.50&38.17& 34.39& 39.98& 41.72& 38.37& 31.53& 40.05\\
    &StkFEP&\textbf{45.25}&\textbf{44.43}& \textbf{48.23}& \textbf{51.57}& \textbf{54.20}& \textbf{40.14}& \textbf{39.57}& \textbf{46.27}\\
    \midrule
    \multirow{4}*{Llama3-8B}&DR + Summ&28.24&38.01& 33.38& 38.66& 39.95& 36.52& \textbf{59.35}& 32.84\\
    &DR + Summ-o-Summ&31.01&35.67& 35.47& 37.61& \textbf{41.89}& 39.60& 46.08& 34.64\\
    &GQR + Summ-o-Summ&35.47&39.07& 34.41& 35.59& 42.30& 42.64& 50.28& 37.54\\
    &StkFEP&\textbf{38.75}&\textbf{39.21}& \textbf{37.61}& \textbf{40.73}& 41.87& \textbf{43.24}& 53.68& \textbf{39.26}\\
 
    \bottomrule
    \end{tabular}
    \label{tab:main_results_cn}
\end{table*}

\noindent \textbf{Similar Event Retrieval. } This module aims to retrieve similar events to reveal potential evolutionary patterns. 
Similar events refer to historically occurred events that are similar to the current question. Similar events can serve as references for predicting future event. For example, in Figure \ref{fig:intro_sample}, by considering news 1, it can be inferred from news 2 whether China will sign a visa exemption agreement with Poland. However, similar events have often been overlooked in previous research. 

Since the extracted stakeholders are mostly specific instances, such as \textit{Australia} and \textit{Li Qiang}, using these stakeholders primarily retrieves relevant events. Retrieving similar events, however, requires more abstract question formulations. To address this, we abstract the stakeholders and then use the abstracted role information to retrieve similar events. 
Specifically, we first use LLM to abstract the stakeholders, obtaining role information such as \textit{country} and \textit{government officials}. Then, based on the original question, background, and stakeholder roles, we generate diverse questions and use these questions to retrieve news about similar events from news sources. Each news article represents a similar event. Since a single news article may not provide comprehensive information, we use the LLM to generate multiple questions to further expand the information about the similar events.

\subsection{Integration}
\label{sec_integration}

The Integration module employs clustering method to clarify the dependencies between events and reduce redundant information. 
After obtaining the relevant and similar events, due to the large scale of retrieved information, models often fail to fully utilize long-range contexts, and performance tends to decrease as context length increases. In addition, models do not rely on all retrieved information, which contains a considerable amount of redundancy. Therefore, before making predictions, it is necessary to clarify the dependencies between events and eliminate redundant information. Prior work often employs summarization methods ~\cite{sarthi2024raptorrecursiveabstractiveprocessing, halawi2024approachinghumanlevelforecastinglanguage}, which generate summaries for each document. However, this method struggles to effectively remove redundancy due to overlapping information among different news  articles.

To address this, we propose a clustering method that organizes text segments into cohesive groups. Specifically, we first extract supportable content segments for prediction from news articles using LLMs and remove any duplicate segments. Next, we cluster all extracted segments. Following existing work ~\cite{guan-etal-2024-tacoere}, we use the K-means clustering algorithm. To determine the optimal number of clusters, we employ the Bayesian information criterion, which not only penalizes model complexity but also rewards goodness of fit ~\cite{adeyemo2024optimal}. After dividing the segments into different clusters, we prompt the LLM to generate a description for each cluster. Finally, these cluster descriptions are utilized to support predictions.

Both relevant and similar events undergo this processing procedure. However, since the outcomes are unknown for predictions, relevant events focus more on gathering comprehensive information, such as in Figure \ref{fig:intro_sample}, the news 2 ``\textit{The President of Poland announced an upcoming visit to China}''. For similar events, where the outcomes are known, the focus is on the outcome and causes of the events, such as retrieved news 1 ``\textit{visits Thailand and signs a mutual exemption agreement}''.

\subsection{Prediction}
\label{sec_prediction}

The Prediction module aims to predict outcomes based on the information gathered about relevant and similar events. 
LLMs employ step-by-step reasoning or self-reflection to enhance their ability to answer questions. However, LLMs often display overconfidence or high randomness~\cite{xiong2024can,yang2024trustllmsmitigateoverconfidence}, frequently providing stubborn or inconsistent feedback~\cite{zhang2024selfcontrastbetterreflectioninconsistent}, which can lead to potential bias and miscalibration. To address this, we implement an aggregate strategy to obtain final prediction outcomes. This strategy involves collecting potential answers from various perspectives of relevant events and similar events, comparing differences between these answers to reach the final result. For time-related questions, the prediction model outputs the time interval with the highest probability. For the remaining questions, the model outputs free-form text.

\begin{table*}[t]
    \centering
        \caption{Model performance of different types of questions on English data (\%).}
    \begin{tabular}{ll|cccccccc}
    \toprule
    Models&Methods&Time&Location&Development&Outcome&Impact&Response&Other&Overall\\
    \midrule
    \multirow{4}*{GPT-3.5}&DR + Summ&35.18&29.98& 32.93& 46.24& 50.51& 35.96& 42.93& 37.41\\
    &DR + Summ-o-Summ&38.12&37.44& 29.21& 49.84& 53.55& 38.74& 48.34& 39.85\\
    &GQR + Summ-o-Summ&42.87&34.65& 33.47& 48.29& 57.59& 45.77& 51.89& 43.58\\
    &StkFEP&\textbf{44.85}&\textbf{38.63}& \textbf{35.81}& \textbf{50.42}& \textbf{60.68}& \textbf{49.74}& \textbf{52.03}& \textbf{46.03}\\
    \midrule
    \multirow{4}*{GLM-4}&DR + Summ&33.53&35.79& 34.32& 39.51& 46.87& 32.37& 32.72& 35.86\\
    &DR + Summ-o-Summ&38.26&30.91& 35.88& 40.81& 47.51& 36.15& \textbf{50.40}& 39.54\\
    &GQR + Summ-o-Summ&42.88&45.47& 33.68& 40.59& 51.24& 40.24& 36.52& 42.62\\
    &StkFEP&\textbf{43.31}&\textbf{48.31}& \textbf{36.40}& \textbf{41.39}& \textbf{54.70}& \textbf{40.63}& 37.57& \textbf{43.11}\\
    \midrule
    \multirow{4}*{Llama3-8B}&DR + Summ&26.34&22.92& 28.24& 48.73& 42.80& 34.11& 25.29& 31.34\\
    &DR + Summ-o-Summ&29.15&23.17& 25.77& 46.82& 44.90& 40.50& 31.96& 32.51\\
    &GQR + Summ-o-Summ&34.50&11.13& 28.18& 38.98& \textbf{55.04}& 38.65& 32.92& 35.16\\
    &StkFEP&\textbf{38.66}&\textbf{29.50}& \textbf{28.35}& \textbf{48.93}& 50.93& \textbf{41.44}& \textbf{46.29}& \textbf{38.77}\\
    \midrule
    \multirow{4}*{Mistral-7B}&DR + Summ&32.26&32.52& 30.48& 35.32& 46.48& 32.83& 41.20& 34.12\\
    &DR + Summ-o-Summ&35.87&31.06& 30.98& 37.25& 54.36& 29.88& 38.06& 36.70\\
    &GQR + Summ-o-Summ&39.12&\textbf{43.01}& 31.56& 36.72&\textbf{ 55.84}& 32.03& 36.81& 38.94\\
    &StkFEP&\textbf{41.38}&32.53&\textbf{ 31.79}&\textbf{ 43.07}& 51.58& \textbf{37.93}& \textbf{57.84}& \textbf{41.24}\\   
    \bottomrule
    \end{tabular}
    \label{tab:main_results_en}
\end{table*}

\section{Experiments}

\subsection{Implementation Details}
\label{exp_implementation_details}

\textbf{Dataset Details. } 
We use GPT-4 to assist in building the dataset. We annotate predictive questions daily, complete annotations the same day, and conduct testing immediately. After the prediction window, we construct ground truth and complete the evaluation. We utilize Bing API to retrieve the news.

\noindent \textbf{Framework Details. } 
We employ multiple advanced LLMs as the backbone, including GPT-3.5\footnote{\url{https://chat.openai.com/chat}}, GLM-4~\cite{glm2024chatglmfamilylargelanguage}, Llama3-8B~\cite{touvron2023llama2openfoundation}, and Mistral-7B~\cite{jiang2023mistral7b}. Due to the limited Chinese data in training Mistral model, it cannot well support Chinese understanding. So Mistral is tested only on English data. We use embedding models, such as Sentence-BERT~\cite{reimers-gurevych-2019-sentence}, to encode the text for clustering.

\noindent \textbf{Evaluation Details. } 
For automatic evaluation by LLM, we use GPT-4 to assess the performance of model predictions. To fully leverage the capabilities of the LLMs, we conduct tests for each dimension separately, such as \textit{Accuracy}, \textit{Completeness}, \textit{Relevance}, \textit{Specificity}, and \textit{Reasonableness}.

\subsection{Baselines}
\label{exp_baseline}

Due to the lack of existing LLM-based methods for open-ended future event prediction, we integrate widely used techniques from different stages to construct the baselines. 

For Retrieval, we select two comparison methods: (1) \textit{DR}, which uses the original predictive question to retrieve information directly; (2) \textit{GQR}, where the LLMs automatically generates multiple questions based on the question or background for retrieval, similar to existing work~\cite{halawi2024approachinghumanlevelforecastinglanguage,cheng2024sociodojo}. 

For Integration, we select two comparison methods: (1) \textit{Summ}, which generates a summary for each retrieved document, similar to existing work~\cite{yan2024autocastenhancingworldevent,halawi2024approachinghumanlevelforecastinglanguage}; (2) \textit{Summ-over-Summ}, which first generates summaries for each document and then produces a brief description of all summaries.  

Finally, for each backbone LLM, we employ three combination strategies as baselines, including \textit{DR + Summ}, \textit{DR + Summ-over-Summ}, and \textit{GQR + Summ-over-Summ}. For the prediction module, all baselines utilize the same prediction framework.

\subsection{Overall Results}
\label{exp_main_results}

The comparative performances of various methods on Chinese and English data are detailed in Tables \ref{tab:main_results_cn} and \ref{tab:main_results_en} respectively. Our approach StkFEP, which integrates stakeholders insights and information from similar events, consistently outperformed other methods. We also have four key observations: 

(1) For time-related questions, the current best result is 44.85\%. In our experiments, we set these questions as multiple-choice format and divided the prediction window into three intervals. Additionally, we tested the performance of GPT-3.5 over five intervals and found a decrease to 25.23\%, indicating that these questions remain highly challenging. 

(2) From the perspective of retrieval methods, using prediction questions directly for retrieval yields the poorest results, while employing LLMs to generate diverse questions shows improvement. 

(3) In terms of information integration, the \textit{Summ-o-Summ} approach, which uses summarization twice, performs better than a single summarization \textit{Summ}, indicating that this method can further refine content.

(4) From the perspective of different languages, the model exhibits similar trends across all languages. The performance on questions related to \textit{Development} is relatively lower.

\subsection{Human Evaluation}
\label{exp_human_evaluation}

\begin{figure}[t]
\centering
\begin{tabular}{cc}
\begin{minipage}[t]{0.45\linewidth}
   \includegraphics[width = 1\linewidth]{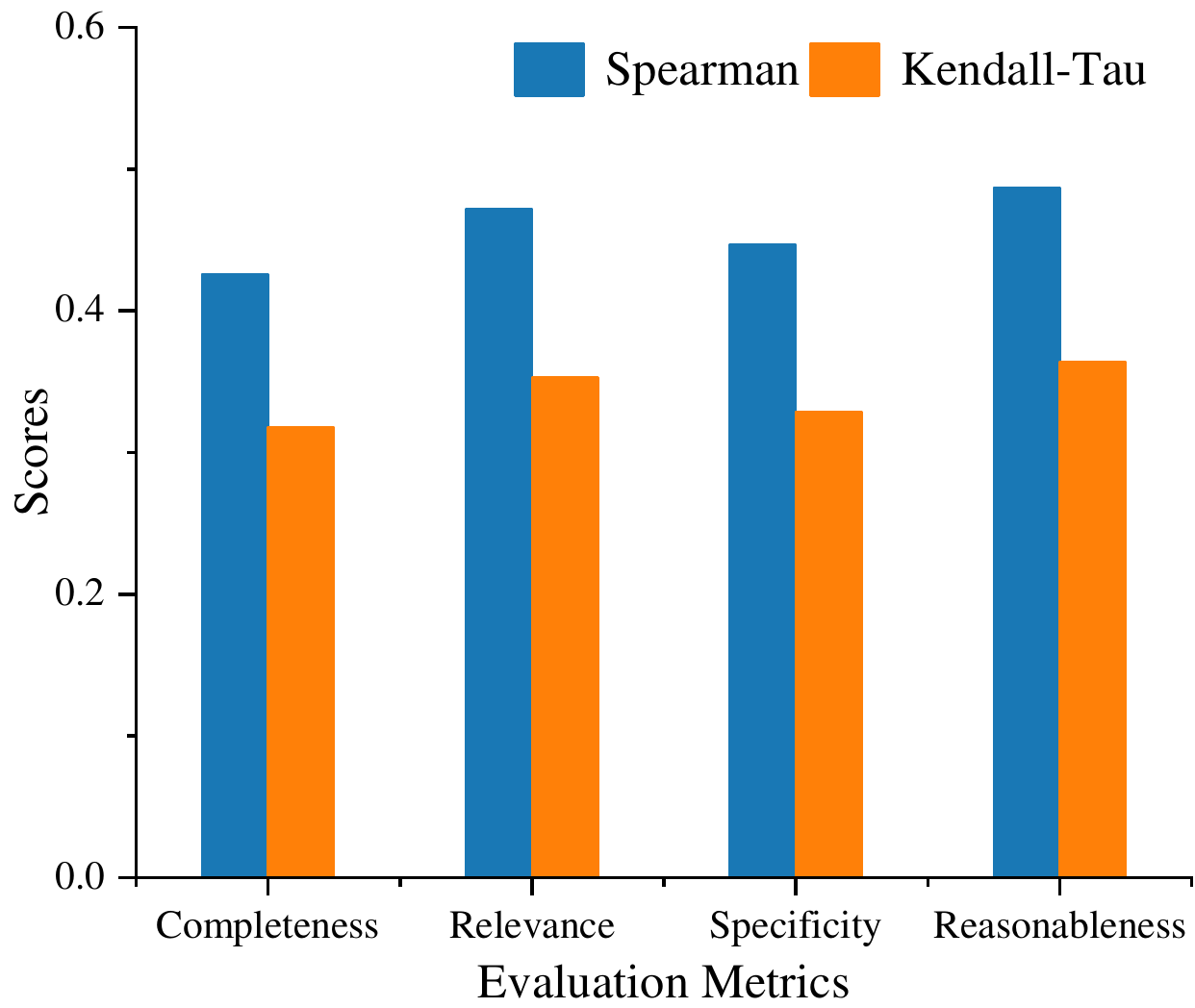}
   \caption{The correlations between human and LLMs.}
   \label{fig:human_evaluation}
\end{minipage}
\hfill
\begin{minipage}[t]{0.45\linewidth}
   \includegraphics[width = 1\linewidth]{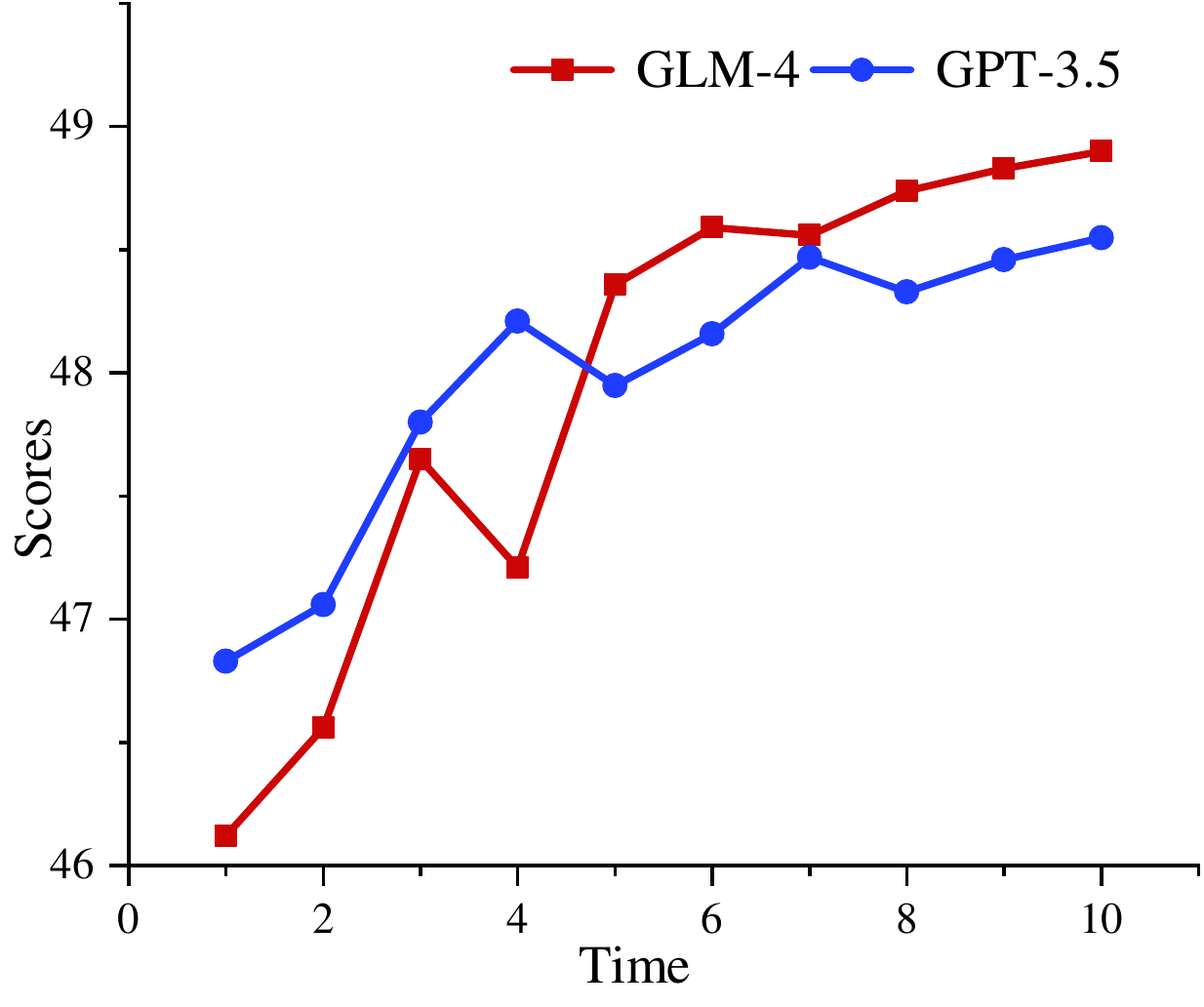}
   \caption{Performance of daily prediction.}
   \label{fig:daily_prediction}
\end{minipage}
\end{tabular}
\end{figure}

In this section, we expand our evaluation methodology beyond model-based metric. We conduct an additional human evaluation to compare 50 predictions generated by GPT-3.5. We invite annotators to assess the model outputs from four dimensions: \textit{Completeness}, \textit{Relevance}, \textit{Specificity}, and \textit{Reasonableness}, using the same criteria as the automatic evaluation method. We report the Spearman and Kendall-Tau correlations between human expert-annotated scores and GPT-4 assigned scores  in Figure \ref{fig:human_evaluation}. We find that GPT-4 achieves a Spearman correlation of around 0.45, which indicates that recent LLMs perform predictions evaluations that are reasonably valid to a meaningful extent.

\subsection{Analysis of Daily Prediction}
\label{exp_phase_pred_analysis}

We conduct daily predictions to capture the trends of predictions changing over time. To achieve this, we select 22 questions that will yield results after 10 days, organizing a test each day. The experimental results, as shown in Figure \ref{fig:daily_prediction}, indicate that the model performance generally improves over time with updates in information. Upon deeper analysis, we observe that during the initial days, the scale of information is substantial, encompassing both redundant and critical details, leading to significant fluctuations. As time progresses, public discussion about the issues diminishes, resulting in smaller fluctuations during this phase.

\begin{figure}[t]
    \centering
    \includegraphics[width=\linewidth]{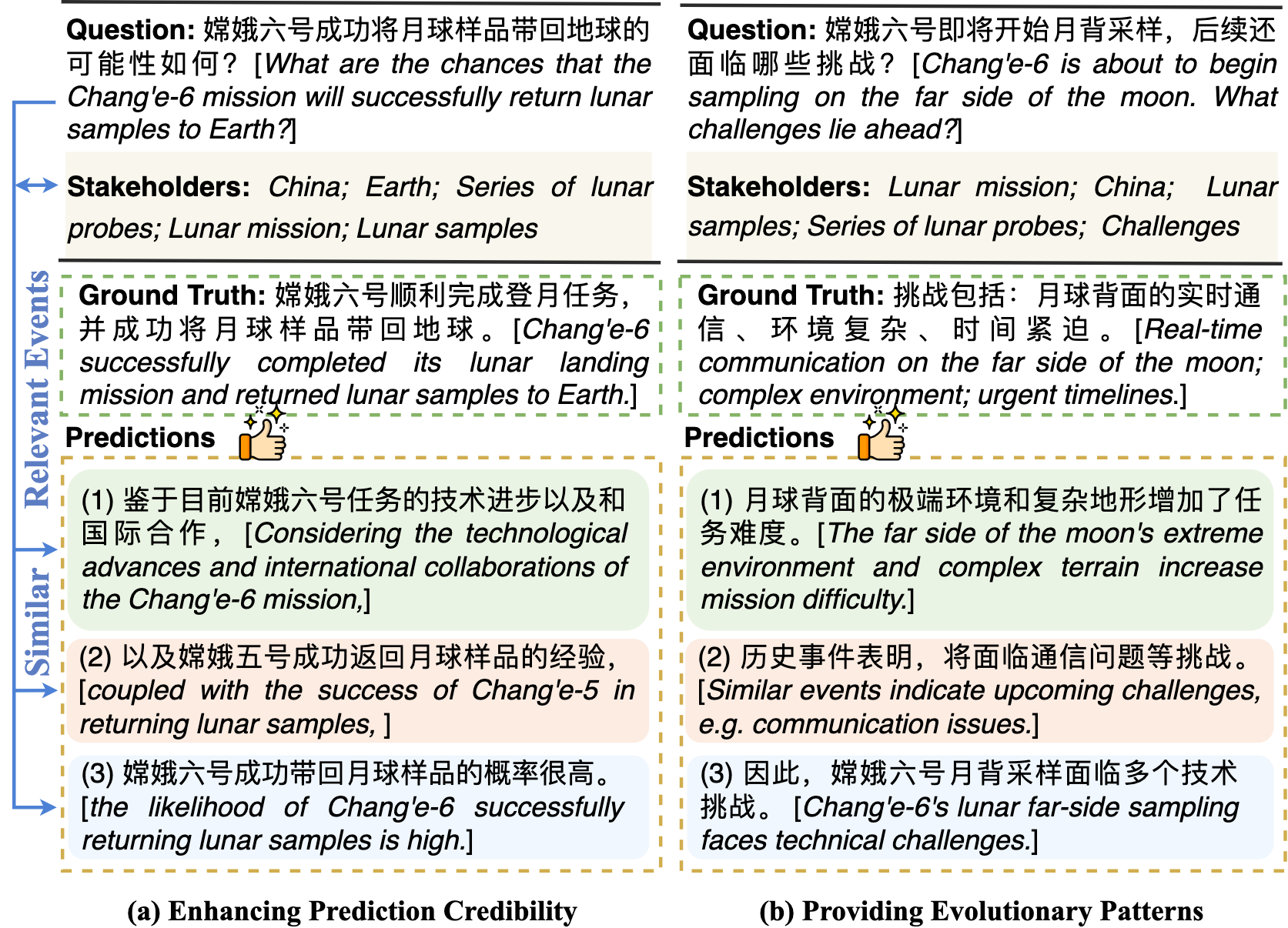}
    \caption{Cases for model predictions.}
    \label{fig:case_study}
\end{figure}

\subsection{Case Study}
\label{exp_case_study}

To better understand the results shown in Table \ref{tab:ablation_study}, we conduct a case study to explicitly illustrate the effectiveness of the event prediction framework. The cases are shown in Figure \ref{fig:case_study}. 
For the first case, by identifying stakeholders such as \textit{Lunar mission}, \textit{Lunar samples}, and \textit{China}, model can effectively retrieve similar events like ``\textit{Chang'e-5 successfully returning lunar samples}''. By then incorporating relevant events, it can significantly enhance the credibility of the predictions. 
For the second case, similar events can provide potential evolutionary patterns to support prediction. Retrieving similar events allows us to learn about challenges faced by previous lunar sampling missions, such as \textit{communication issues}, and combining this with the progress and breakthroughs in current research, can enhance the effectiveness of event prediction.

\subsection{Error Analysis}
\label{exp_error_analysis}

\begin{figure}[t]
    \centering
    \includegraphics[width=\linewidth]{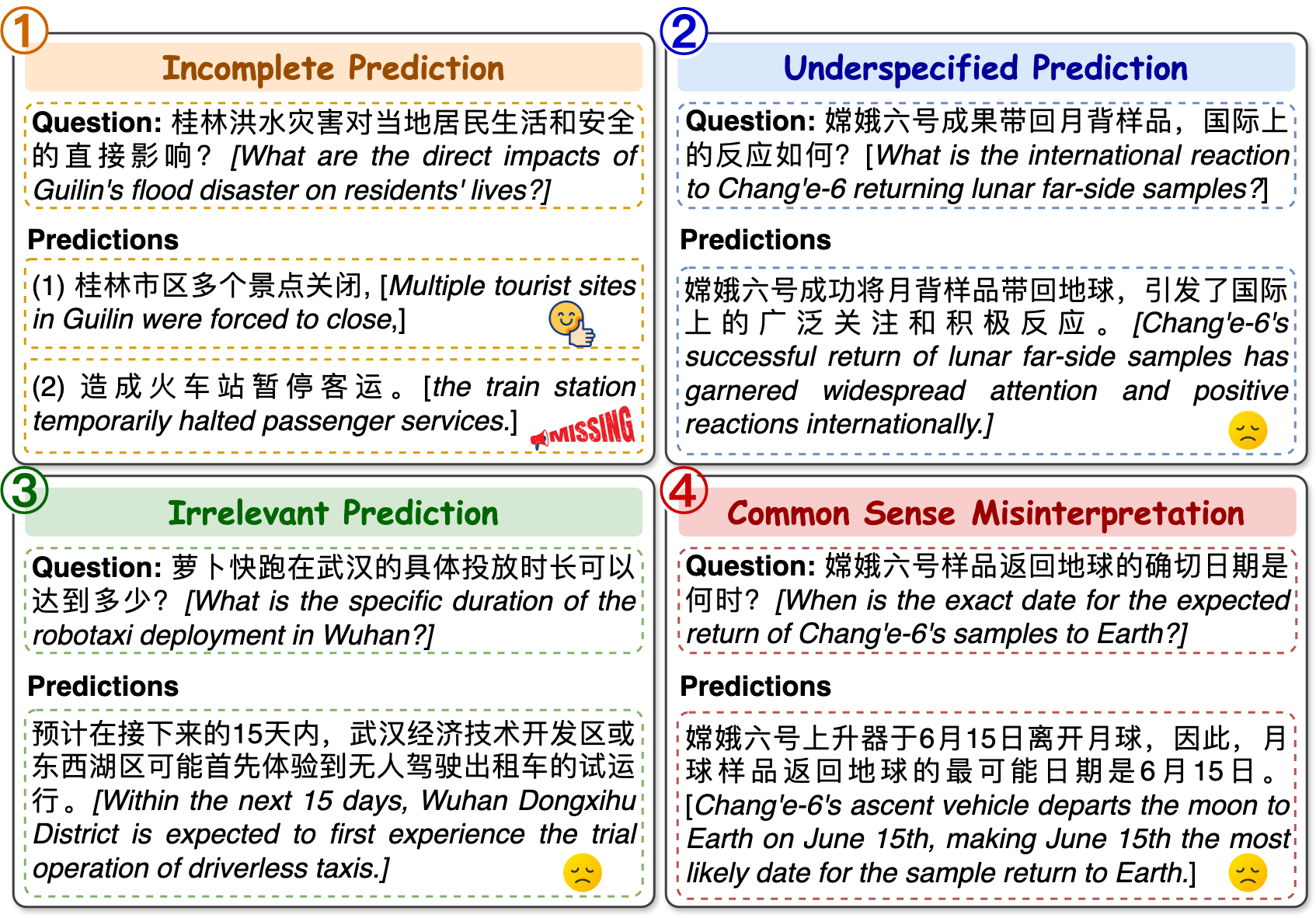}
    \caption{Error analysis of the model predictions.}
    \label{fig:error_analysis}
\end{figure}

To enrich the understanding and better advance future research, we conduct a detailed analysis of the problems encountered in existing research. The common problems can primarily be divided into four categories: 
(1) \textbf{Incomplete Prediction } refers to scenarios where the predictions made are not comprehensive enough to cover all aspects or variables related to the event. As shown in case 1 of Figure \ref{fig:error_analysis}, the model overlooks the outcome ``\textit{the train station temporarily halted passenger services}''. 
(2) \textbf{Underspecified Prediction } occurs when predictions are too vague or general, lacking specific details necessary for them to be actionable or useful. As shown in case 2, the model outputs ``\textit{Chang'e-6's successful return of lunar far-side samples has garnered widespread attention and positive reactions internationally}''. The predictions of model lacks value because it does not provide any salient entity information, resulting in an output too generic to effectively address the specific question. 
(3) \textbf{Irrelevant Prediction } describes predictions is unrelated to the question posed, essentially providing answers that do not address the question. As shown in case 3, the question asks about time information, but the model responds with a location information ``\textit{Wuhan Dongxihu District}''. 
(4) \textbf{Common Sense Misinterpretation } arises when predictions contradict basic common sense, resulting in outcomes that are implausible or logically inconsistent with known facts. This undermines the credibility of the predictions and may lead to mistrust or disregard of model outputs. In case 4, the statement ``\textit{Chang'e-6's ascent vehicle departs the moon for Earth on June 15th}'' is predicted, however, the model overlooks the common sense that it is impossible to return from the moon to Earth within a day.

\begin{table}[h]
    \centering
    \caption{Ablation study.}    
    \begin{tabular}{lccc}
    \toprule
         Methods& GPT-3.5& GLM-4  \\
         \midrule
         StkFEP& 46.95&46.27\\
         w/o Cluster-Summ& 46.11&45.38\\
         w/o Similar Events& 45.65&44.79\\
         w/o Stakeholders& 44.28&42.80\\
         \bottomrule
    \end{tabular}
    \label{tab:ablation_study}
\end{table}

\subsection{Ablation Study}
\label{exp_main_results}

To more specifically validate the different modules within the event prediction framework, we conduct experiments to ablate the clustering-over-summarization method (\textit{w/o Cluster-Summ}) for information integration, similar events (\textit{w/o Similar Events}), and stakeholders (\textit{w/o Stakeholders}). From the results in Table \ref{tab:ablation_study}, we can see that: (1) For the scenario without cluster summarization (\textit{w/o Cluster-Summ}), where we used Summ-over-Summ for information integration, the model performance decreased, indicating that our method can more effectively refine information and organize dependencies between events.
(2) For the scenario without similar events (\textit{w/o Similar Events}), relying only on relevant events for predictions, the model results also declined, mainly because similar events provide potential evolutionary patterns that support the final predictions.
(3) For the scenario without stakeholders (\textit{w/o Stakeholders}), ignoring stakeholders resulted in the most substantial drop in model performance. This demonstrates that utilizing stakeholders not only enhances the diversity of retrieval but also enables more accurate retrieval of similar events.

\section{Conclusions}

In this paper, we introduce OpenEP (an open-ended future event prediction task), which generates flexible and diverse predictions aligned with real-world scenarios. To facilitate the study of this task, we first construct OpenEPBench, an open-ended future event prediction dataset. For question construction, we pose questions from seven perspectives, including location, time, event development, event outcome, event impact, event response, and other, to facilitate an in-depth analysis and understanding of the comprehensive evolution of events. For outcome construction, we collect free-form text containing the outcomes as ground truth to provide semantically complete and detail-enriched outcomes. Furthermore, we propose StkFEP, a stakeholder-enhanced future event prediction framework that incorporates the characteristics of event evolution for open-ended settings. Our method extracts stakeholders involved in events to extend questions and collects historical events that are relevant and similar to the question to gather diverse and comprehensive information to support model prediction.  Extensive experiments on Chinese and English data demonstrate that accurately predicting future events in open-ended settings is challenging for existing large language models.

\bibliographystyle{ACM-Reference-Format}
\bibliography{sample-base}


\begin{thebibliography}{35}


\ifx \showCODEN    \undefined \def \showCODEN     #1{\unskip}     \fi
\ifx \showDOI      \undefined \def \showDOI       #1{#1}\fi
\ifx \showISBNx    \undefined \def \showISBNx     #1{\unskip}     \fi
\ifx \showISBNxiii \undefined \def \showISBNxiii  #1{\unskip}     \fi
\ifx \showISSN     \undefined \def \showISSN      #1{\unskip}     \fi
\ifx \showLCCN     \undefined \def \showLCCN      #1{\unskip}     \fi
\ifx \shownote     \undefined \def \shownote      #1{#1}          \fi
\ifx \showarticletitle \undefined \def \showarticletitle #1{#1}   \fi
\ifx \showURL      \undefined \def \showURL       {\relax}        \fi
\providecommand\bibfield[2]{#2}
\providecommand\bibinfo[2]{#2}
\providecommand\natexlab[1]{#1}
\providecommand\showeprint[2][]{arXiv:#2}

\bibitem[Adeyemo and Bhattacharyya(2024)]%
        {adeyemo2024optimal}
\bibfield{author}{\bibinfo{person}{Samuel Adeyemo} {and} \bibinfo{person}{Debangsu Bhattacharyya}.} \bibinfo{year}{2024}\natexlab{}.
\newblock \showarticletitle{Optimal nonlinear dynamic sparse model selection and Bayesian parameter estimation for nonlinear systems}.
\newblock \bibinfo{journal}{\emph{Computers \& Chemical Engineering}}  \bibinfo{volume}{180} (\bibinfo{year}{2024}), \bibinfo{pages}{108502}.
\newblock


\bibitem[Bai et~al\mbox{.}(2023)]%
        {bai_rich_aaai_2023}
\bibfield{author}{\bibinfo{person}{Long Bai}, \bibinfo{person}{Saiping Guan}, \bibinfo{person}{Zixuan Li}, \bibinfo{person}{Jiafeng Guo}, \bibinfo{person}{Xiaolong Jin}, {and} \bibinfo{person}{Xueqi Cheng}.} \bibinfo{year}{2023}\natexlab{}.
\newblock \showarticletitle{Rich event modeling for script event prediction}. In \bibinfo{booktitle}{\emph{Proceedings of the AAAI}}. Article \bibinfo{articleno}{1409}, \bibinfo{numpages}{9}~pages.
\newblock


\bibitem[Cheng and Chin(2024)]%
        {cheng2024sociodojo}
\bibfield{author}{\bibinfo{person}{Junyan Cheng} {and} \bibinfo{person}{Peter Chin}.} \bibinfo{year}{2024}\natexlab{}.
\newblock \showarticletitle{SocioDojo: Building Lifelong Analytical Agents with Real-world Text and Time Series}. In \bibinfo{booktitle}{\emph{Proceedings of the ICLR}}.
\newblock


\bibitem[Dempsey et~al\mbox{.}(2017)]%
        {pmlr-v70-dempsey17a}
\bibfield{author}{\bibinfo{person}{Walter~H. Dempsey}, \bibinfo{person}{Alexander Moreno}, \bibinfo{person}{Christy~K. Scott}, \bibinfo{person}{Michael~L. Dennis}, \bibinfo{person}{David~H. Gustafson}, \bibinfo{person}{Susan~A. Murphy}, {and} \bibinfo{person}{James~M. Rehg}.} \bibinfo{year}{2017}\natexlab{}.
\newblock \showarticletitle{i{S}urvive: An Interpretable, Event-time Prediction Model for m{H}ealth}. In \bibinfo{booktitle}{\emph{Proceedings of the ICML}} \emph{(\bibinfo{series}{Proceedings of Machine Learning Research}, Vol.~\bibinfo{volume}{70})}. \bibinfo{pages}{970--979}.
\newblock


\bibitem[GLM et~al\mbox{.}(2024)]%
        {glm2024chatglmfamilylargelanguage}
\bibfield{author}{\bibinfo{person}{Team GLM}, \bibinfo{person}{:}, \bibinfo{person}{Aohan Zeng}, \bibinfo{person}{Bin Xu}, \bibinfo{person}{Bowen Wang}, \bibinfo{person}{Chenhui Zhang}, \bibinfo{person}{Da Yin}, \bibinfo{person}{Diego Rojas}, \bibinfo{person}{Guanyu Feng}, \bibinfo{person}{Hanlin Zhao}, \bibinfo{person}{Hanyu Lai}, \bibinfo{person}{Hao Yu}, \bibinfo{person}{Hongning Wang}, \bibinfo{person}{Jiadai Sun}, \bibinfo{person}{Jiajie Zhang}, \bibinfo{person}{Jiale Cheng}, \bibinfo{person}{Jiayi Gui}, \bibinfo{person}{Jie Tang}, \bibinfo{person}{Jing Zhang}, \bibinfo{person}{Juanzi Li}, \bibinfo{person}{Lei Zhao}, \bibinfo{person}{Lindong Wu}, \bibinfo{person}{Lucen Zhong}, \bibinfo{person}{Mingdao Liu}, \bibinfo{person}{Minlie Huang}, \bibinfo{person}{Peng Zhang}, \bibinfo{person}{Qinkai Zheng}, \bibinfo{person}{Rui Lu}, \bibinfo{person}{Shuaiqi Duan}, \bibinfo{person}{Shudan Zhang}, \bibinfo{person}{Shulin Cao}, \bibinfo{person}{Shuxun Yang}, \bibinfo{person}{Weng~Lam Tam}, \bibinfo{person}{Wenyi
  Zhao}, \bibinfo{person}{Xiao Liu}, \bibinfo{person}{Xiao Xia}, \bibinfo{person}{Xiaohan Zhang}, \bibinfo{person}{Xiaotao Gu}, \bibinfo{person}{Xin Lv}, \bibinfo{person}{Xinghan Liu}, \bibinfo{person}{Xinyi Liu}, \bibinfo{person}{Xinyue Yang}, \bibinfo{person}{Xixuan Song}, \bibinfo{person}{Xunkai Zhang}, \bibinfo{person}{Yifan An}, \bibinfo{person}{Yifan Xu}, \bibinfo{person}{Yilin Niu}, \bibinfo{person}{Yuantao Yang}, \bibinfo{person}{Yueyan Li}, \bibinfo{person}{Yushi Bai}, \bibinfo{person}{Yuxiao Dong}, \bibinfo{person}{Zehan Qi}, \bibinfo{person}{Zhaoyu Wang}, \bibinfo{person}{Zhen Yang}, \bibinfo{person}{Zhengxiao Du}, \bibinfo{person}{Zhenyu Hou}, {and} \bibinfo{person}{Zihan Wang}.} \bibinfo{year}{2024}\natexlab{}.
\newblock \bibinfo{title}{ChatGLM: A Family of Large Language Models from GLM-130B to GLM-4 All Tools}.
\newblock
\newblock
\showeprint[arxiv]{2406.12793}~[cs.CL]


\bibitem[Guan et~al\mbox{.}(2024)]%
        {guan-etal-2024-tacoere}
\bibfield{author}{\bibinfo{person}{Yong Guan}, \bibinfo{person}{Xiaozhi Wang}, \bibinfo{person}{Lei Hou}, \bibinfo{person}{Juanzi Li}, \bibinfo{person}{Jeff~Z. Pan}, \bibinfo{person}{Jiaoyan Chen}, {and} \bibinfo{person}{Freddy Lecue}.} \bibinfo{year}{2024}\natexlab{}.
\newblock \showarticletitle{{T}aco{ERE}: Cluster-aware Compression for Event Relation Extraction}. In \bibinfo{booktitle}{\emph{Proceedings of the LREC-COLING}}. \bibinfo{pages}{15511--15521}.
\newblock


\bibitem[Halawi et~al\mbox{.}(2024)]%
        {halawi2024approachinghumanlevelforecastinglanguage}
\bibfield{author}{\bibinfo{person}{Danny Halawi}, \bibinfo{person}{Fred Zhang}, \bibinfo{person}{Chen Yueh-Han}, {and} \bibinfo{person}{Jacob Steinhardt}.} \bibinfo{year}{2024}\natexlab{}.
\newblock \bibinfo{title}{Approaching Human-Level Forecasting with Language Models}.
\newblock
\newblock
\showeprint[arxiv]{2402.18563}~[cs.LG]


\bibitem[Hashimoto et~al\mbox{.}(2014)]%
        {hashimoto-etal-2014-toward}
\bibfield{author}{\bibinfo{person}{Chikara Hashimoto}, \bibinfo{person}{Kentaro Torisawa}, \bibinfo{person}{Julien Kloetzer}, \bibinfo{person}{Motoki Sano}, \bibinfo{person}{Istv{\'a}n Varga}, \bibinfo{person}{Jong-Hoon Oh}, {and} \bibinfo{person}{Yutaka Kidawara}.} \bibinfo{year}{2014}\natexlab{}.
\newblock \showarticletitle{Toward Future Scenario Generation: Extracting Event Causality Exploiting Semantic Relation, Context, and Association Features}. In \bibinfo{booktitle}{\emph{Proceedings of the ACL}}. \bibinfo{pages}{987--997}.
\newblock


\bibitem[Jiang et~al\mbox{.}(2023)]%
        {jiang2023mistral7b}
\bibfield{author}{\bibinfo{person}{Albert~Q. Jiang}, \bibinfo{person}{Alexandre Sablayrolles}, \bibinfo{person}{Arthur Mensch}, \bibinfo{person}{Chris Bamford}, \bibinfo{person}{Devendra~Singh Chaplot}, \bibinfo{person}{Diego de~las Casas}, \bibinfo{person}{Florian Bressand}, \bibinfo{person}{Gianna Lengyel}, \bibinfo{person}{Guillaume Lample}, \bibinfo{person}{Lucile Saulnier}, \bibinfo{person}{Lélio~Renard Lavaud}, \bibinfo{person}{Marie-Anne Lachaux}, \bibinfo{person}{Pierre Stock}, \bibinfo{person}{Teven~Le Scao}, \bibinfo{person}{Thibaut Lavril}, \bibinfo{person}{Thomas Wang}, \bibinfo{person}{Timothée Lacroix}, {and} \bibinfo{person}{William~El Sayed}.} \bibinfo{year}{2023}\natexlab{}.
\newblock \bibinfo{title}{Mistral 7B}.
\newblock
\newblock
\showeprint[arxiv]{2310.06825}~[cs.CL]


\bibitem[Kuila and Sarkar(2024)]%
        {kuila_stak_sigir_2024}
\bibfield{author}{\bibinfo{person}{Alapan Kuila} {and} \bibinfo{person}{Sudeshna Sarkar}.} \bibinfo{year}{2024}\natexlab{}.
\newblock \showarticletitle{From Text to Context: An Entailment Approach for News Stakeholder Classification}. In \bibinfo{booktitle}{\emph{Proceedings of the SIGIR}}. \bibinfo{pages}{2426–2430}.
\newblock
\showISBNx{9798400704314}


\bibitem[Kwak et~al\mbox{.}(2010)]%
        {kwak2010twitter}
\bibfield{author}{\bibinfo{person}{Haewoon Kwak}, \bibinfo{person}{Changhyun Lee}, \bibinfo{person}{Hosung Park}, {and} \bibinfo{person}{Sue Moon}.} \bibinfo{year}{2010}\natexlab{}.
\newblock \showarticletitle{What is Twitter, a social network or a news media?}. In \bibinfo{booktitle}{\emph{Proceedings of the 19th international conference on World wide web}}. \bibinfo{pages}{591--600}.
\newblock


\bibitem[Laxman et~al\mbox{.}(2008)]%
        {laxman_stream_kdd_2008}
\bibfield{author}{\bibinfo{person}{Srivatsan Laxman}, \bibinfo{person}{Vikram Tankasali}, {and} \bibinfo{person}{Ryen~W. White}.} \bibinfo{year}{2008}\natexlab{}.
\newblock \showarticletitle{Stream prediction using a generative model based on frequent episodes in event sequences}. In \bibinfo{booktitle}{\emph{Proceedings of the SIGKDD}}. \bibinfo{pages}{453–461}.
\newblock


\bibitem[Li et~al\mbox{.}(2021)]%
        {li-etal-2021-future}
\bibfield{author}{\bibinfo{person}{Manling Li}, \bibinfo{person}{Sha Li}, \bibinfo{person}{Zhenhailong Wang}, \bibinfo{person}{Lifu Huang}, \bibinfo{person}{Kyunghyun Cho}, \bibinfo{person}{Heng Ji}, \bibinfo{person}{Jiawei Han}, {and} \bibinfo{person}{Clare Voss}.} \bibinfo{year}{2021}\natexlab{}.
\newblock \showarticletitle{The Future is not One-dimensional: Complex Event Schema Induction by Graph Modeling for Event Prediction}. In \bibinfo{booktitle}{\emph{Proceedings of the EMNLP}}, \bibfield{editor}{\bibinfo{person}{Marie-Francine Moens}, \bibinfo{person}{Xuanjing Huang}, \bibinfo{person}{Lucia Specia}, {and} \bibinfo{person}{Scott Wen-tau Yih}} (Eds.). \bibinfo{pages}{5203--5215}.
\newblock


\bibitem[Liu et~al\mbox{.}(2022)]%
        {liu2022event}
\bibfield{author}{\bibinfo{person}{Xiaoyan Liu}, \bibinfo{person}{Jiarui Zhao}, \bibinfo{person}{Ran Liu}, {and} \bibinfo{person}{Kai Liu}.} \bibinfo{year}{2022}\natexlab{}.
\newblock \showarticletitle{Event history analysis of the duration of online public opinions regarding major health emergencies}.
\newblock \bibinfo{journal}{\emph{Frontiers in Psychology}}  \bibinfo{volume}{13} (\bibinfo{year}{2022}), \bibinfo{pages}{954559}.
\newblock


\bibitem[Ma et~al\mbox{.}(2023a)]%
        {ma_forcast_kdd_2023}
\bibfield{author}{\bibinfo{person}{Yunshan Ma}, \bibinfo{person}{Chenchen Ye}, \bibinfo{person}{Zijian Wu}, \bibinfo{person}{Xiang Wang}, \bibinfo{person}{Yixin Cao}, {and} \bibinfo{person}{Tat-Seng Chua}.} \bibinfo{year}{2023}\natexlab{a}.
\newblock \showarticletitle{Context-aware Event Forecasting via Graph Disentanglement}. In \bibinfo{booktitle}{\emph{Proceedings of the SIGKDD}}. \bibinfo{pages}{1643–1652}.
\newblock


\bibitem[Ma et~al\mbox{.}(2023b)]%
        {ma_context_kdd_2023}
\bibfield{author}{\bibinfo{person}{Yunshan Ma}, \bibinfo{person}{Chenchen Ye}, \bibinfo{person}{Zijian Wu}, \bibinfo{person}{Xiang Wang}, \bibinfo{person}{Yixin Cao}, {and} \bibinfo{person}{Tat-Seng Chua}.} \bibinfo{year}{2023}\natexlab{b}.
\newblock \showarticletitle{Context-aware Event Forecasting via Graph Disentanglement}. In \bibinfo{booktitle}{\emph{Proceedings of the SIGKDD}}. \bibinfo{pages}{1643–1652}.
\newblock


\bibitem[Narsupalli et~al\mbox{.}(2024)]%
        {narsupalli2024reviewfeedbackreasonrefernovelframework}
\bibfield{author}{\bibinfo{person}{Yaswanth Narsupalli}, \bibinfo{person}{Abhranil Chandra}, \bibinfo{person}{Sreevatsa Muppirala}, \bibinfo{person}{Manish Gupta}, {and} \bibinfo{person}{Pawan Goyal}.} \bibinfo{year}{2024}\natexlab{}.
\newblock \bibinfo{title}{Review-Feedback-Reason (ReFeR): A Novel Framework for NLG Evaluation and Reasoning}.
\newblock
\newblock
\showeprint[arxiv]{2407.12877}~[cs.CL]


\bibitem[OpenAI(2023)]%
        {openai2023gpt4}
\bibfield{author}{\bibinfo{person}{OpenAI}.} \bibinfo{year}{2023}\natexlab{}.
\newblock \bibinfo{title}{GPT-4 Technical Report}.
\newblock
\newblock
\showeprint[arxiv]{2303.08774}
\urldef\tempurl%
\url{https://arxiv.org/pdf/2303.08774.pdf}
\showURL{%
\tempurl}


\bibitem[Pratt et~al\mbox{.}(2024)]%
        {pratt2024languagemodelsuseforecasting}
\bibfield{author}{\bibinfo{person}{Sarah Pratt}, \bibinfo{person}{Seth Blumberg}, \bibinfo{person}{Pietro~Kreitlon Carolino}, {and} \bibinfo{person}{Meredith~Ringel Morris}.} \bibinfo{year}{2024}\natexlab{}.
\newblock \bibinfo{title}{Can Language Models Use Forecasting Strategies?}
\newblock
\newblock
\showeprint[arxiv]{2406.04446}~[cs.LG]


\bibitem[Reimers and Gurevych(2019)]%
        {reimers-gurevych-2019-sentence}
\bibfield{author}{\bibinfo{person}{Nils Reimers} {and} \bibinfo{person}{Iryna Gurevych}.} \bibinfo{year}{2019}\natexlab{}.
\newblock \showarticletitle{Sentence-{BERT}: Sentence Embeddings using {S}iamese {BERT}-Networks}. In \bibinfo{booktitle}{\emph{Proceedings of the EMNLP-IJCNLP}}, \bibfield{editor}{\bibinfo{person}{Kentaro Inui}, \bibinfo{person}{Jing Jiang}, \bibinfo{person}{Vincent Ng}, {and} \bibinfo{person}{Xiaojun Wan}} (Eds.). \bibinfo{pages}{3982--3992}.
\newblock


\bibitem[Rumi et~al\mbox{.}(2018)]%
        {pumi_pred_sigspatial_2018}
\bibfield{author}{\bibinfo{person}{Shakila~Khan Rumi}, \bibinfo{person}{Ke Deng}, {and} \bibinfo{person}{Flora~D. Salim}.} \bibinfo{year}{2018}\natexlab{}.
\newblock \showarticletitle{Theft prediction with individual risk factor of visitors}. In \bibinfo{booktitle}{\emph{Proceedings of the SIGSPATIAL}}. \bibinfo{pages}{552–555}.
\newblock


\bibitem[Sarthi et~al\mbox{.}(2024)]%
        {sarthi2024raptorrecursiveabstractiveprocessing}
\bibfield{author}{\bibinfo{person}{Parth Sarthi}, \bibinfo{person}{Salman Abdullah}, \bibinfo{person}{Aditi Tuli}, \bibinfo{person}{Shubh Khanna}, \bibinfo{person}{Anna Goldie}, {and} \bibinfo{person}{Christopher~D. Manning}.} \bibinfo{year}{2024}\natexlab{}.
\newblock \bibinfo{title}{RAPTOR: Recursive Abstractive Processing for Tree-Organized Retrieval}.
\newblock
\newblock
\showeprint[arxiv]{2401.18059}~[cs.CL]


\bibitem[Schoenegger et~al\mbox{.}(2024)]%
        {schoenegger2024wisdomsiliconcrowdllm}
\bibfield{author}{\bibinfo{person}{Philipp Schoenegger}, \bibinfo{person}{Indre Tuminauskaite}, \bibinfo{person}{Peter~S. Park}, \bibinfo{person}{Rafael Valdece~Sousa Bastos}, {and} \bibinfo{person}{Philip~E. Tetlock}.} \bibinfo{year}{2024}\natexlab{}.
\newblock \bibinfo{title}{Wisdom of the Silicon Crowd: LLM Ensemble Prediction Capabilities Rival Human Crowd Accuracy}.
\newblock
\newblock
\showeprint[arxiv]{2402.19379}~[cs.CY]


\bibitem[Sharma et~al\mbox{.}(2013)]%
        {sharma2013news}
\bibfield{author}{\bibinfo{person}{Smriti Sharma}, \bibinfo{person}{Rajesh Kumar}, \bibinfo{person}{Pawan Bhadana}, {and} \bibinfo{person}{Sumita Gupta}.} \bibinfo{year}{2013}\natexlab{}.
\newblock \showarticletitle{News event extraction using 5W1H approach \& its analysis}.
\newblock \bibinfo{journal}{\emph{International Journal of Scientific \& Engineering Research}} \bibinfo{volume}{4}, \bibinfo{number}{5} (\bibinfo{year}{2013}), \bibinfo{pages}{2064--2068}.
\newblock


\bibitem[Touvron et~al\mbox{.}(2023)]%
        {touvron2023llama2openfoundation}
\bibfield{author}{\bibinfo{person}{Hugo Touvron}, \bibinfo{person}{Louis Martin}, \bibinfo{person}{Kevin Stone}, \bibinfo{person}{Peter Albert}, \bibinfo{person}{Amjad Almahairi}, \bibinfo{person}{Yasmine Babaei}, \bibinfo{person}{Nikolay Bashlykov}, \bibinfo{person}{Soumya Batra}, \bibinfo{person}{Prajjwal Bhargava}, \bibinfo{person}{Shruti Bhosale}, \bibinfo{person}{Dan Bikel}, \bibinfo{person}{Lukas Blecher}, \bibinfo{person}{Cristian~Canton Ferrer}, \bibinfo{person}{Moya Chen}, \bibinfo{person}{Guillem Cucurull}, \bibinfo{person}{David Esiobu}, \bibinfo{person}{Jude Fernandes}, \bibinfo{person}{Jeremy Fu}, \bibinfo{person}{Wenyin Fu}, \bibinfo{person}{Brian Fuller}, \bibinfo{person}{Cynthia Gao}, \bibinfo{person}{Vedanuj Goswami}, \bibinfo{person}{Naman Goyal}, \bibinfo{person}{Anthony Hartshorn}, \bibinfo{person}{Saghar Hosseini}, \bibinfo{person}{Rui Hou}, \bibinfo{person}{Hakan Inan}, \bibinfo{person}{Marcin Kardas}, \bibinfo{person}{Viktor Kerkez}, \bibinfo{person}{Madian Khabsa},
  \bibinfo{person}{Isabel Kloumann}, \bibinfo{person}{Artem Korenev}, \bibinfo{person}{Punit~Singh Koura}, \bibinfo{person}{Marie-Anne Lachaux}, \bibinfo{person}{Thibaut Lavril}, \bibinfo{person}{Jenya Lee}, \bibinfo{person}{Diana Liskovich}, \bibinfo{person}{Yinghai Lu}, \bibinfo{person}{Yuning Mao}, \bibinfo{person}{Xavier Martinet}, \bibinfo{person}{Todor Mihaylov}, \bibinfo{person}{Pushkar Mishra}, \bibinfo{person}{Igor Molybog}, \bibinfo{person}{Yixin Nie}, \bibinfo{person}{Andrew Poulton}, \bibinfo{person}{Jeremy Reizenstein}, \bibinfo{person}{Rashi Rungta}, \bibinfo{person}{Kalyan Saladi}, \bibinfo{person}{Alan Schelten}, \bibinfo{person}{Ruan Silva}, \bibinfo{person}{Eric~Michael Smith}, \bibinfo{person}{Ranjan Subramanian}, \bibinfo{person}{Xiaoqing~Ellen Tan}, \bibinfo{person}{Binh Tang}, \bibinfo{person}{Ross Taylor}, \bibinfo{person}{Adina Williams}, \bibinfo{person}{Jian~Xiang Kuan}, \bibinfo{person}{Puxin Xu}, \bibinfo{person}{Zheng Yan}, \bibinfo{person}{Iliyan Zarov}, \bibinfo{person}{Yuchen
  Zhang}, \bibinfo{person}{Angela Fan}, \bibinfo{person}{Melanie Kambadur}, \bibinfo{person}{Sharan Narang}, \bibinfo{person}{Aurelien Rodriguez}, \bibinfo{person}{Robert Stojnic}, \bibinfo{person}{Sergey Edunov}, {and} \bibinfo{person}{Thomas Scialom}.} \bibinfo{year}{2023}\natexlab{}.
\newblock \bibinfo{title}{Llama 2: Open Foundation and Fine-Tuned Chat Models}.
\newblock
\newblock
\showeprint[arxiv]{2307.09288}~[cs.CL]


\bibitem[Xiong et~al\mbox{.}(2024)]%
        {xiong2024can}
\bibfield{author}{\bibinfo{person}{Miao Xiong}, \bibinfo{person}{Zhiyuan Hu}, \bibinfo{person}{Xinyang Lu}, \bibinfo{person}{YIFEI LI}, \bibinfo{person}{Jie Fu}, \bibinfo{person}{Junxian He}, {and} \bibinfo{person}{Bryan Hooi}.} \bibinfo{year}{2024}\natexlab{}.
\newblock \showarticletitle{Can {LLM}s Express Their Uncertainty? An Empirical Evaluation of Confidence Elicitation in {LLM}s}. In \bibinfo{booktitle}{\emph{Proceedings of the ICLR}}.
\newblock


\bibitem[Yan et~al\mbox{.}(2024)]%
        {yan2024autocastenhancingworldevent}
\bibfield{author}{\bibinfo{person}{Qi Yan}, \bibinfo{person}{Raihan Seraj}, \bibinfo{person}{Jiawei He}, \bibinfo{person}{Lili Meng}, {and} \bibinfo{person}{Tristan Sylvain}.} \bibinfo{year}{2024}\natexlab{}.
\newblock \bibinfo{title}{AutoCast++: Enhancing World Event Prediction with Zero-shot Ranking-based Context Retrieval}.
\newblock
\newblock
\showeprint[arxiv]{2310.01880}~[cs.LG]


\bibitem[Yang et~al\mbox{.}(2024)]%
        {yang2024trustllmsmitigateoverconfidence}
\bibfield{author}{\bibinfo{person}{Haoyan Yang}, \bibinfo{person}{Yixuan Wang}, \bibinfo{person}{Xingyin Xu}, \bibinfo{person}{Hanyuan Zhang}, {and} \bibinfo{person}{Yirong Bian}.} \bibinfo{year}{2024}\natexlab{}.
\newblock \bibinfo{title}{Can We Trust LLMs? Mitigate Overconfidence Bias in LLMs through Knowledge Transfer}.
\newblock
\newblock
\showeprint[arxiv]{2405.16856}~[cs.CL]


\bibitem[Yang et~al\mbox{.}(2019)]%
        {yang_financial_gnn_cikm_2019}
\bibfield{author}{\bibinfo{person}{Yiying Yang}, \bibinfo{person}{Zhongyu Wei}, \bibinfo{person}{Qin Chen}, {and} \bibinfo{person}{Libo Wu}.} \bibinfo{year}{2019}\natexlab{}.
\newblock \showarticletitle{Using External Knowledge for Financial Event Prediction Based on Graph Neural Networks}. In \bibinfo{booktitle}{\emph{Proceedings of the CIKM}}. \bibinfo{pages}{2161–2164}.
\newblock


\bibitem[Ye et~al\mbox{.}(2024)]%
        {ye2024miraievaluatingllmagents}
\bibfield{author}{\bibinfo{person}{Chenchen Ye}, \bibinfo{person}{Ziniu Hu}, \bibinfo{person}{Yihe Deng}, \bibinfo{person}{Zijie Huang}, \bibinfo{person}{Mingyu~Derek Ma}, \bibinfo{person}{Yanqiao Zhu}, {and} \bibinfo{person}{Wei Wang}.} \bibinfo{year}{2024}\natexlab{}.
\newblock \bibinfo{title}{MIRAI: Evaluating LLM Agents for Event Forecasting}.
\newblock
\newblock
\showeprint[arxiv]{2407.01231}~[cs.CL]


\bibitem[Zhang et~al\mbox{.}(2024b)]%
        {zhang2024selfcontrastbetterreflectioninconsistent}
\bibfield{author}{\bibinfo{person}{Wenqi Zhang}, \bibinfo{person}{Yongliang Shen}, \bibinfo{person}{Linjuan Wu}, \bibinfo{person}{Qiuying Peng}, \bibinfo{person}{Jun Wang}, \bibinfo{person}{Yueting Zhuang}, {and} \bibinfo{person}{Weiming Lu}.} \bibinfo{year}{2024}\natexlab{b}.
\newblock \bibinfo{title}{Self-Contrast: Better Reflection Through Inconsistent Solving Perspectives}.
\newblock
\newblock
\showeprint[arxiv]{2401.02009}~[cs.CL]


\bibitem[Zhang et~al\mbox{.}(2024a)]%
        {zhang2024largelanguagemodelsevaluators}
\bibfield{author}{\bibinfo{person}{Xiaoyu Zhang}, \bibinfo{person}{Yishan Li}, \bibinfo{person}{Jiayin Wang}, \bibinfo{person}{Bowen Sun}, \bibinfo{person}{Weizhi Ma}, \bibinfo{person}{Peijie Sun}, {and} \bibinfo{person}{Min Zhang}.} \bibinfo{year}{2024}\natexlab{a}.
\newblock \bibinfo{title}{Large Language Models as Evaluators for Recommendation Explanations}.
\newblock
\newblock
\showeprint[arxiv]{2406.03248}~[cs.IR]


\bibitem[Zhao(2021)]%
        {zhao2021event}
\bibfield{author}{\bibinfo{person}{Liang Zhao}.} \bibinfo{year}{2021}\natexlab{}.
\newblock \showarticletitle{Event prediction in the big data era: A systematic survey}.
\newblock \bibinfo{journal}{\emph{ACM Computing Surveys (CSUR)}} \bibinfo{volume}{54}, \bibinfo{number}{5} (\bibinfo{year}{2021}), \bibinfo{pages}{1--37}.
\newblock


\bibitem[Zhu et~al\mbox{.}(2023)]%
        {zhu_generative_aaai_2023}
\bibfield{author}{\bibinfo{person}{Fangqi Zhu}, \bibinfo{person}{Jun Gao}, \bibinfo{person}{Changlong Yu}, \bibinfo{person}{Wei Wang}, \bibinfo{person}{Chen Xu}, \bibinfo{person}{Xin Mu}, \bibinfo{person}{Min Yang}, {and} \bibinfo{person}{Ruifeng Xu}.} \bibinfo{year}{2023}\natexlab{}.
\newblock \showarticletitle{A generative approach for script event prediction via contrastive fine-tuning}. In \bibinfo{booktitle}{\emph{Proceedings of the AAAI}}. Article \bibinfo{articleno}{1576}, \bibinfo{numpages}{9}~pages.
\newblock


\bibitem[Zou et~al\mbox{.}(2022)]%
        {NEURIPS2022_aec870a6}
\bibfield{author}{\bibinfo{person}{Andy Zou}, \bibinfo{person}{Tristan Xiao}, \bibinfo{person}{Ryan Jia}, \bibinfo{person}{Joe Kwon}, \bibinfo{person}{Mantas Mazeika}, \bibinfo{person}{Richard Li}, \bibinfo{person}{Dawn Song}, \bibinfo{person}{Jacob Steinhardt}, \bibinfo{person}{Owain Evans}, {and} \bibinfo{person}{Dan Hendrycks}.} \bibinfo{year}{2022}\natexlab{}.
\newblock \showarticletitle{Forecasting Future World Events With Neural Networks}. In \bibinfo{booktitle}{\emph{Proceedings of the NeurIPS}}, Vol.~\bibinfo{volume}{35}. \bibinfo{pages}{27293--27305}.
\newblock


\end{thebibliography}

\appendix
\clearpage

\begin{figure}
    \centering
    \includegraphics[width=0.8\linewidth]{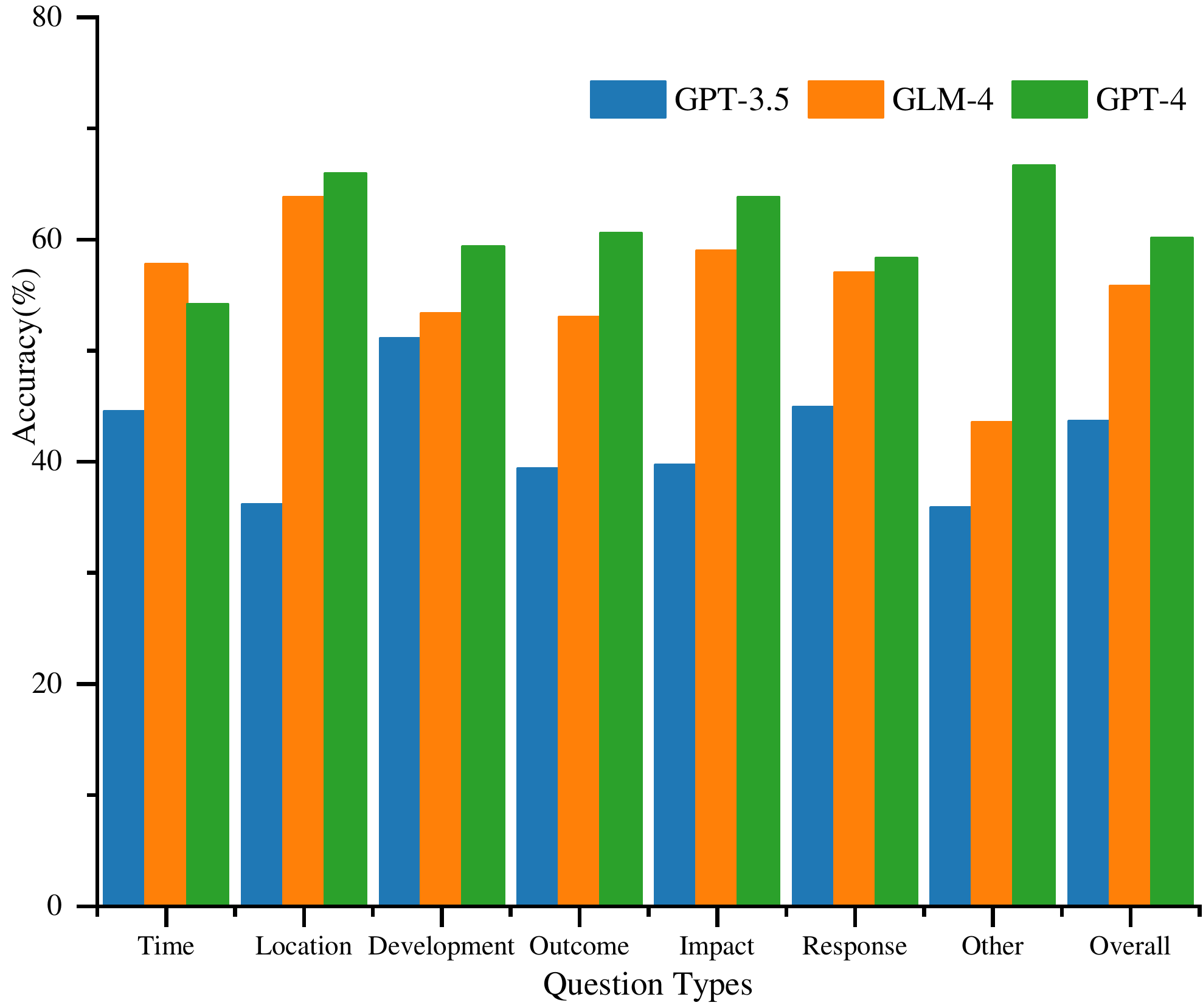}
    \caption{Performance of data validity verification on different LLMs.}
    \label{fig:validity_check}
\end{figure}

\section{Data Validity Verification}
\label{appendix_data_valid_verify}

We adopt a data construction strategy that involves annotating questions on the same day they are tested, with answers collected after the prediction window for evaluation. This means that at the time of question annotation, it is unknown whether there will be outcomes, inevitably leading to instances where no answers are available. Such instances are termed as invalid questions, and are excluded from the dataset. Notably, we collect 983 valid questions and annotate 286 invalid ones. Although these questions lack answers, they are valuable for evaluating the model's capacity to identify question validity. Therefore, based on these 286 invalid questions, we extract an equivalent number of valid questions to test whether three powerful LLMs, such as GPT-4, GLM-4, and GPT-3.5, can identify which questions would have answers and which would not during the prediction window.

The results on different types of questions are shown in Figure \ref{fig:validity_check}. The results indicate that event identify the data validity is challenging for existing LLMs. We also have the following three observations: 
(1) For model perspectives, GPT-4 achieves the best performance, except in time-related questions where GLM-4 slightly outperforms. 
(2) Regarding question types, time-related questions exhibit the lowest accuracy, and GLM-4 achieves the highest accuracy in this category with 57.83\%. 
(3) In terms of overall scores, GPT-4 leads in performance. However, GPT-3.5 achieves an accuracy score of 43.67\%.

\section{Performance on Individual Dimension}
\label{appendix_indi_dim}

\begin{figure*}[t]
\centering
\begin{tabular}{cc}
\begin{minipage}[t]{0.28\linewidth}
   \includegraphics[width = 1\linewidth]{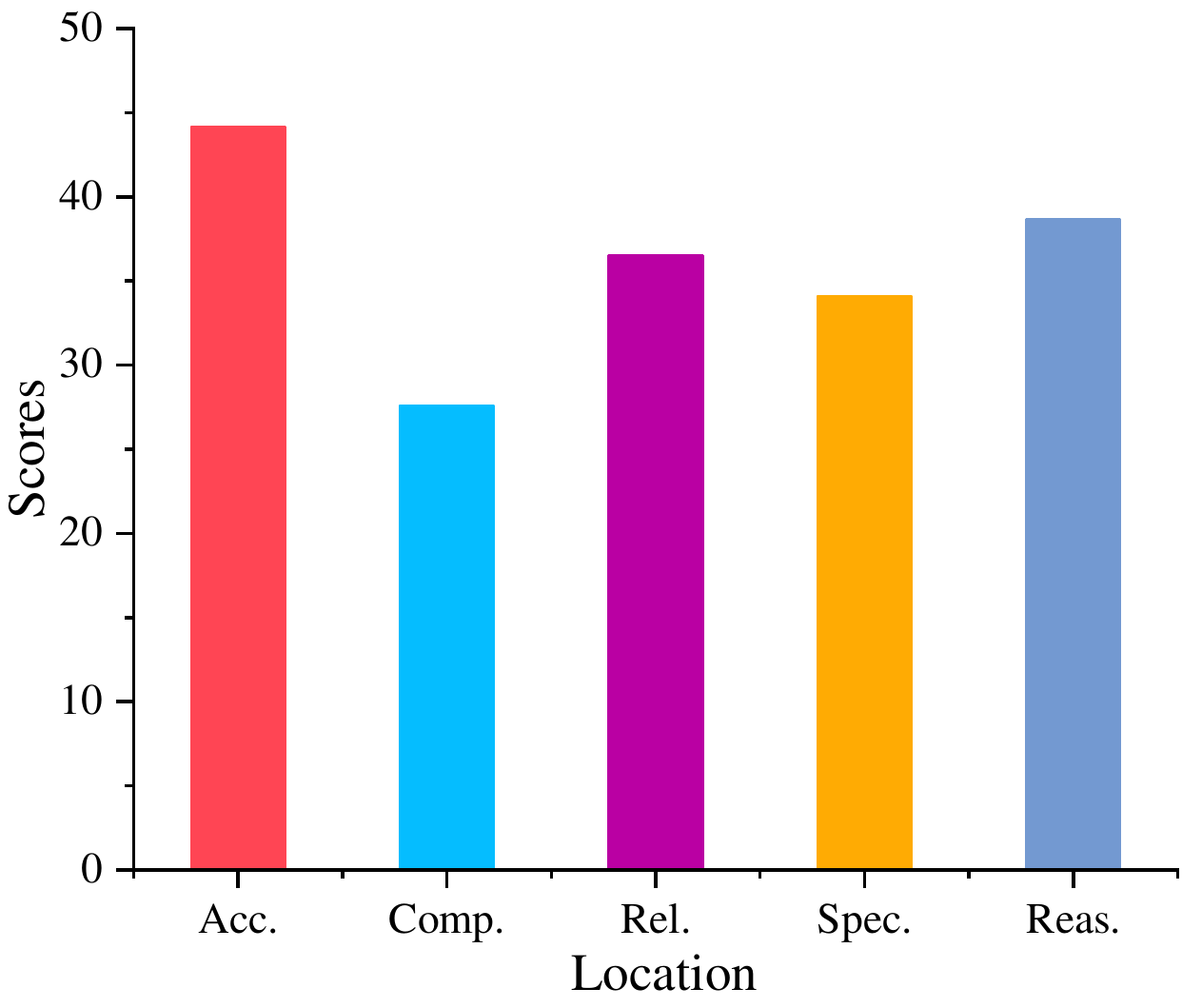}
\end{minipage}
\hfill
\begin{minipage}[t]{0.28\linewidth}
   \includegraphics[width = 1\linewidth]{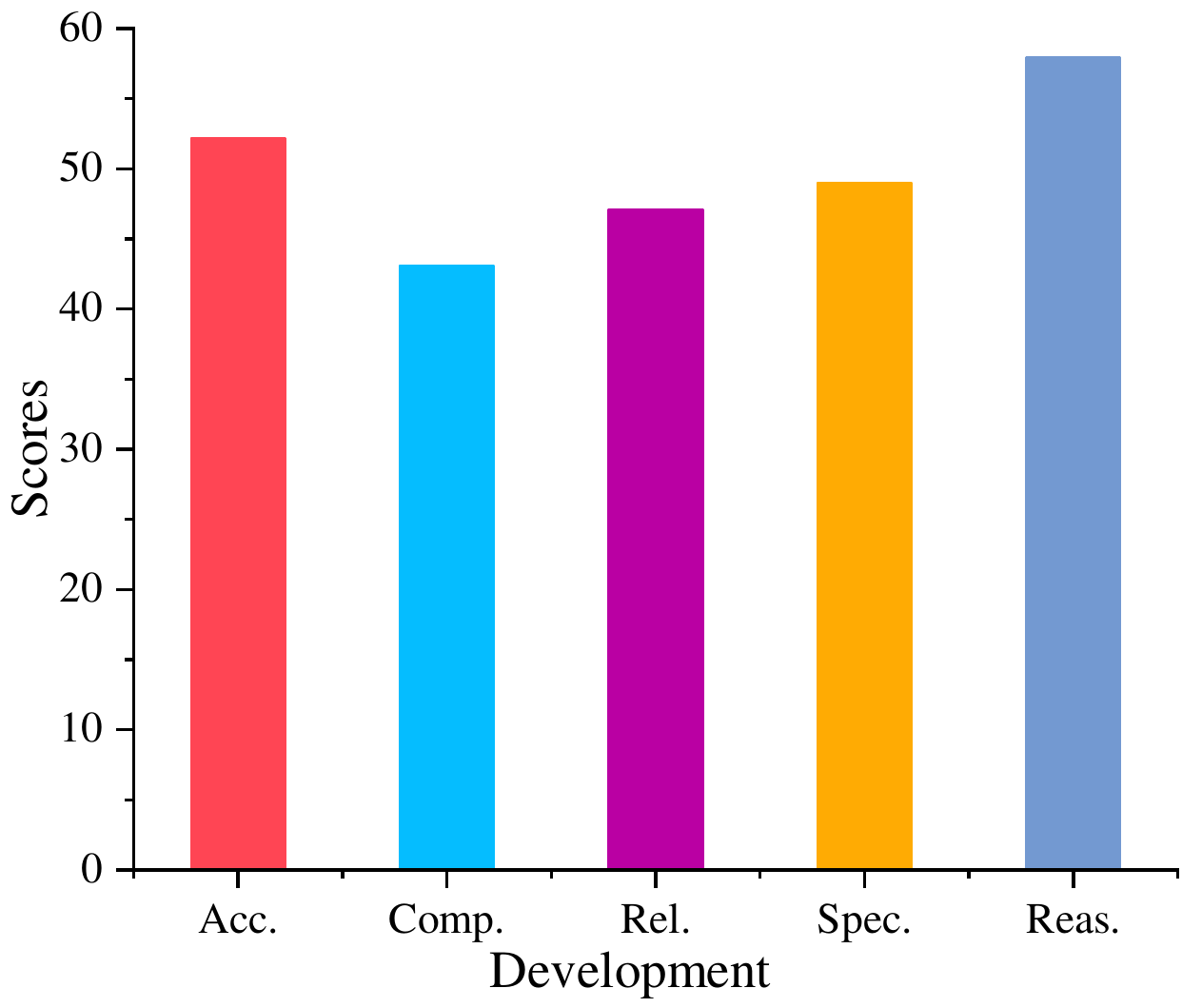}
\end{minipage}
\hfill
\begin{minipage}[t]{0.28\linewidth}
   \includegraphics[width = 1\linewidth]{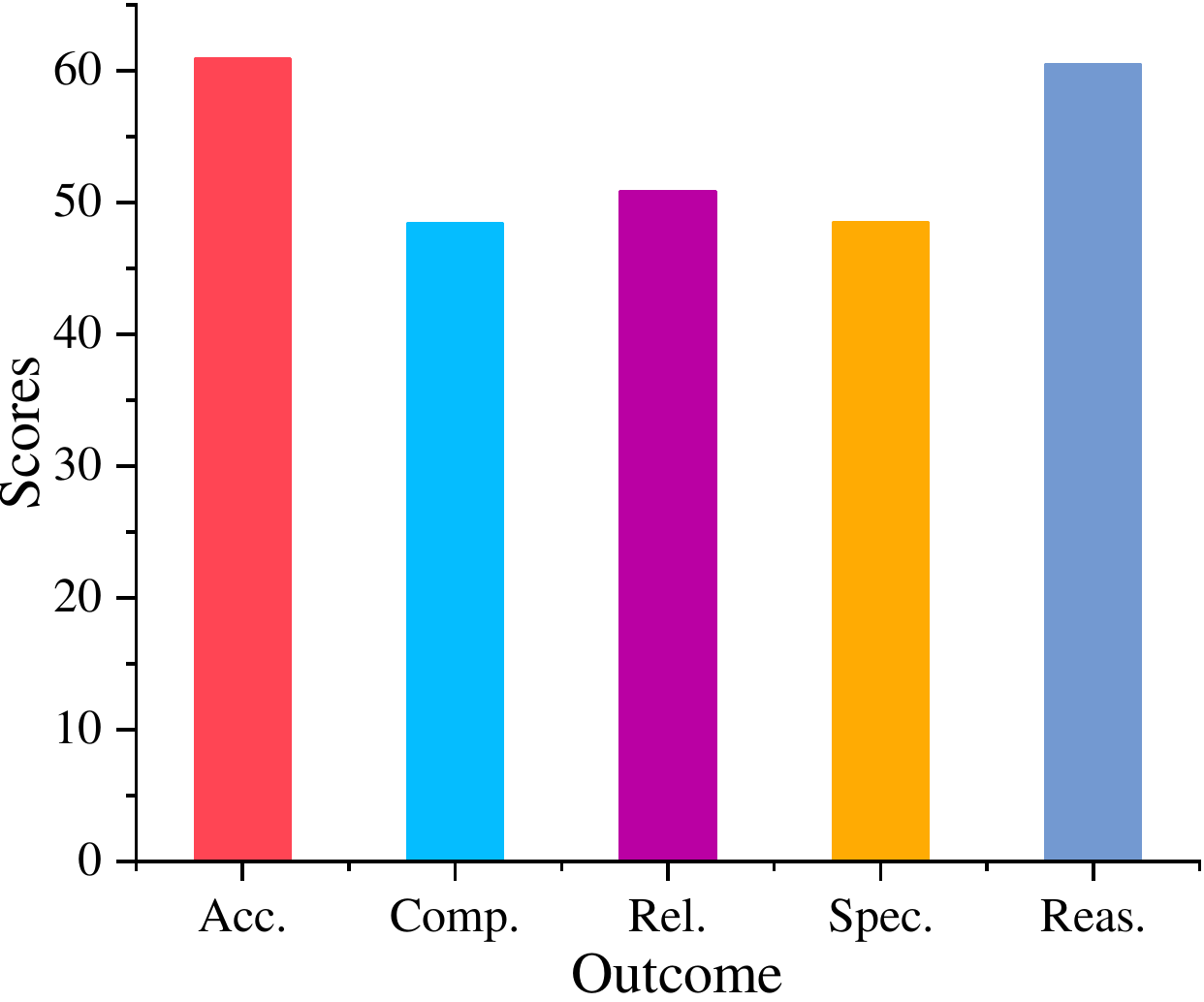}
\end{minipage}
\end{tabular}
\caption{GPT-3.5 Performance of individual dimension on Location, Event Development, and Event Outcome. Acc., Comp., Rel., Spec., and Reas. are abbreviations for Accuracy, Completeness, Relevance, Specificity, and Reasonableness, respectively.}
\label{fig:chat_indi_dem_1}
\end{figure*}

\begin{figure*}[t]
\centering
\begin{tabular}{cc}
\begin{minipage}[t]{0.28\linewidth}
   \includegraphics[width = 1\linewidth]{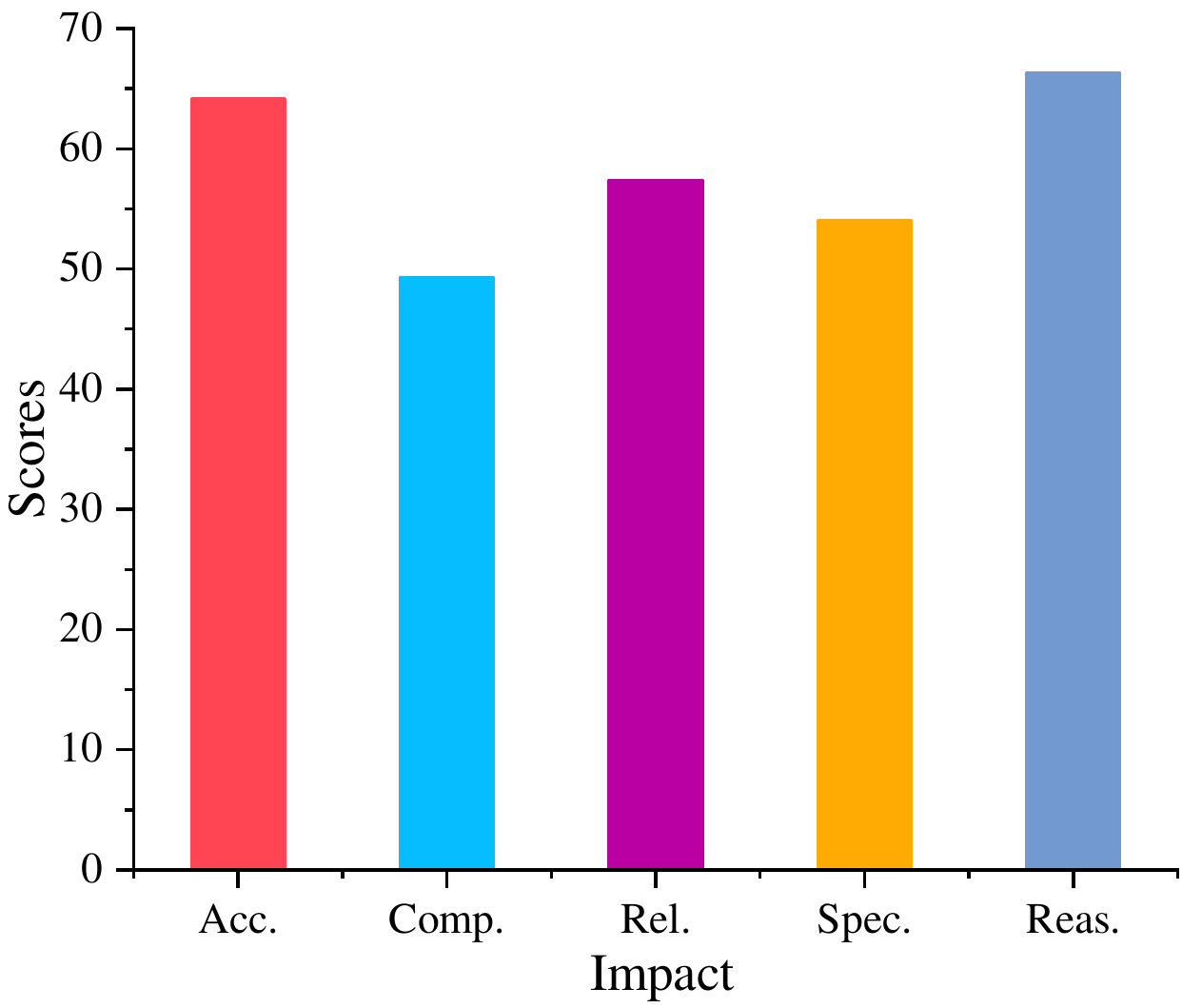}
\end{minipage}
\hfill
\begin{minipage}[t]{0.28\linewidth}
   \includegraphics[width = 1\linewidth]{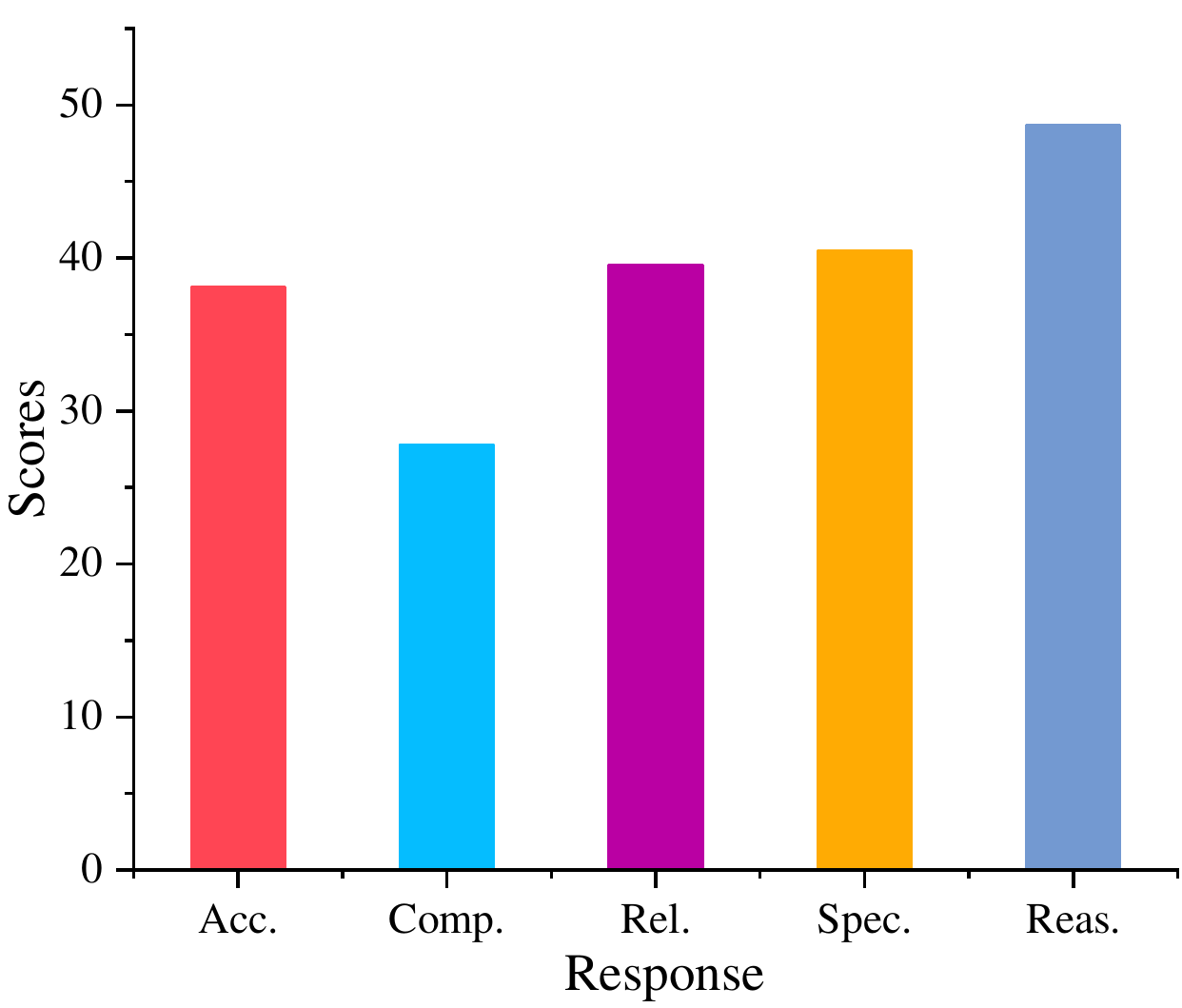}
\end{minipage}
\hfill
\begin{minipage}[t]{0.28\linewidth}
   \includegraphics[width = 1\linewidth]{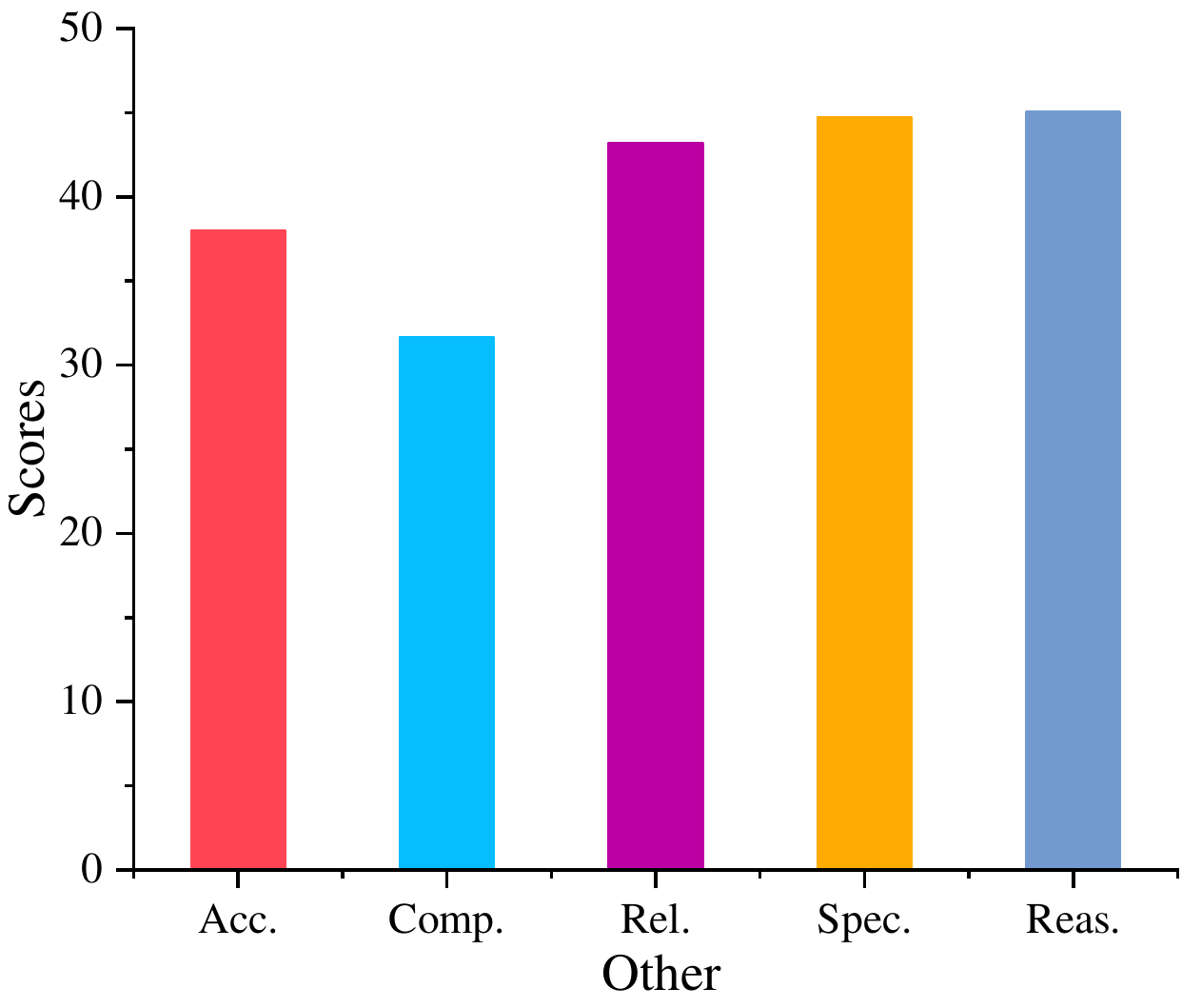}
\end{minipage}
\end{tabular}
\caption{GPT-3.5 Performance of individual dimension on Event Impact, Event Response, and Other.}
\label{fig:chat_indi_dem_2}
\end{figure*}

\begin{figure*}[t]
\centering
\begin{tabular}{cc}
\begin{minipage}[t]{0.28\linewidth}
   \includegraphics[width = 1\linewidth]{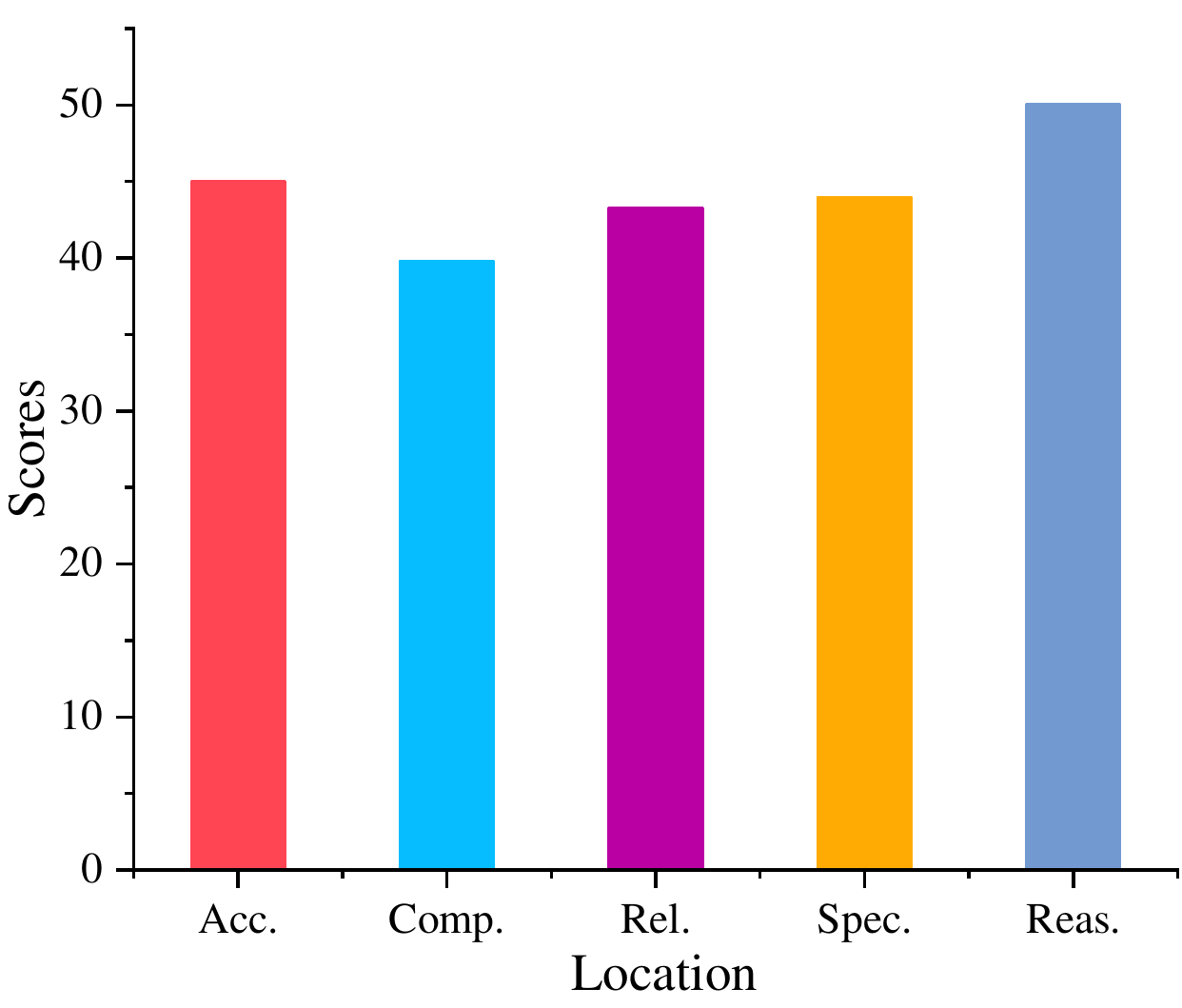}
\end{minipage}
\hfill
\begin{minipage}[t]{0.28\linewidth}
   \includegraphics[width = 1\linewidth]{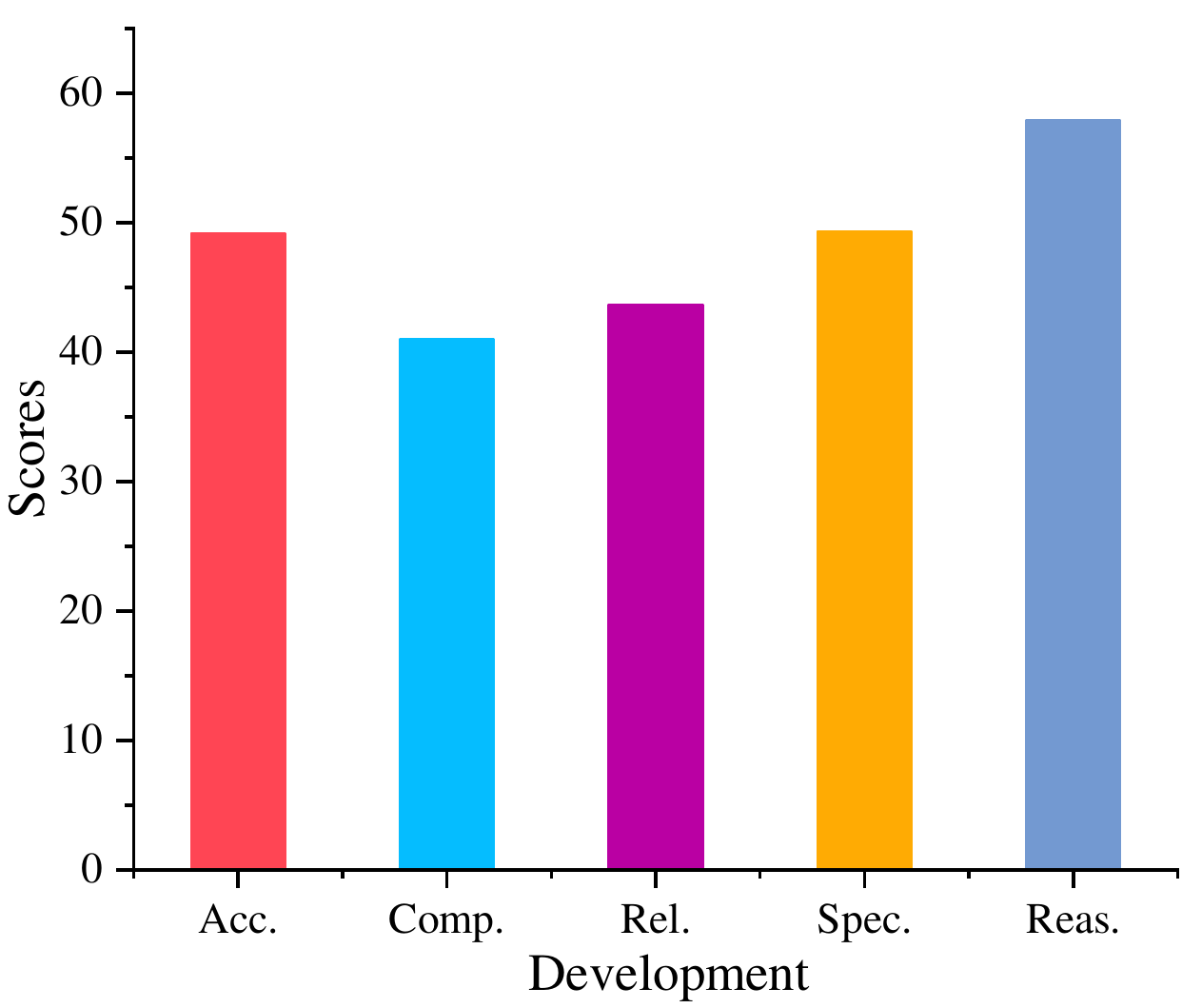}
\end{minipage}
\hfill
\begin{minipage}[t]{0.28\linewidth}
   \includegraphics[width = 1\linewidth]{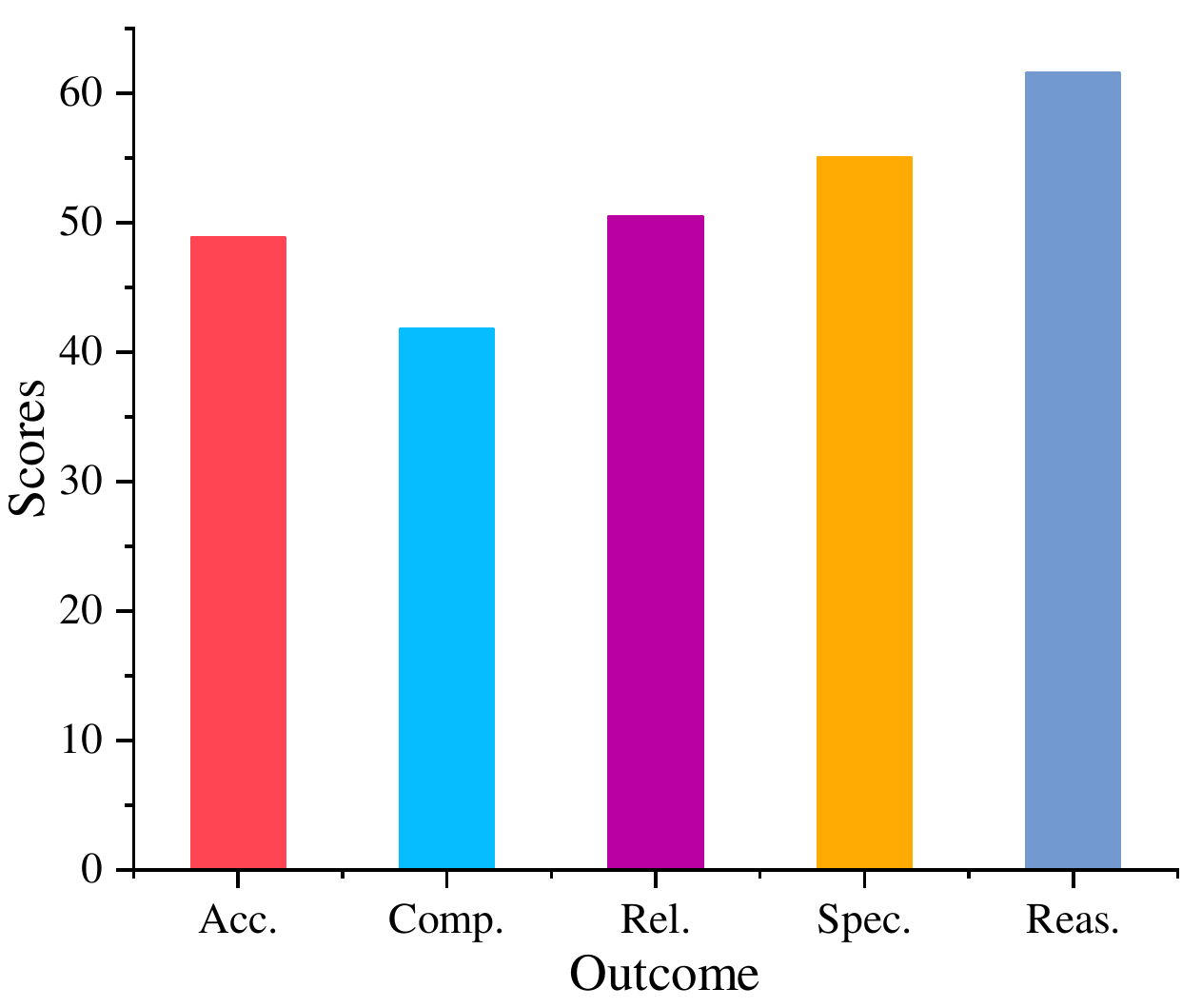}
\end{minipage}
\end{tabular}
\caption{GLM-4 Performance of individual dimension on Location, Event Development, and Event Outcome.}
\label{fig:glm_indi_dim_1}
\end{figure*}

\begin{figure*}[t]
\centering
\begin{tabular}{cc}
\begin{minipage}[t]{0.28\linewidth}
   \includegraphics[width = 1\linewidth]{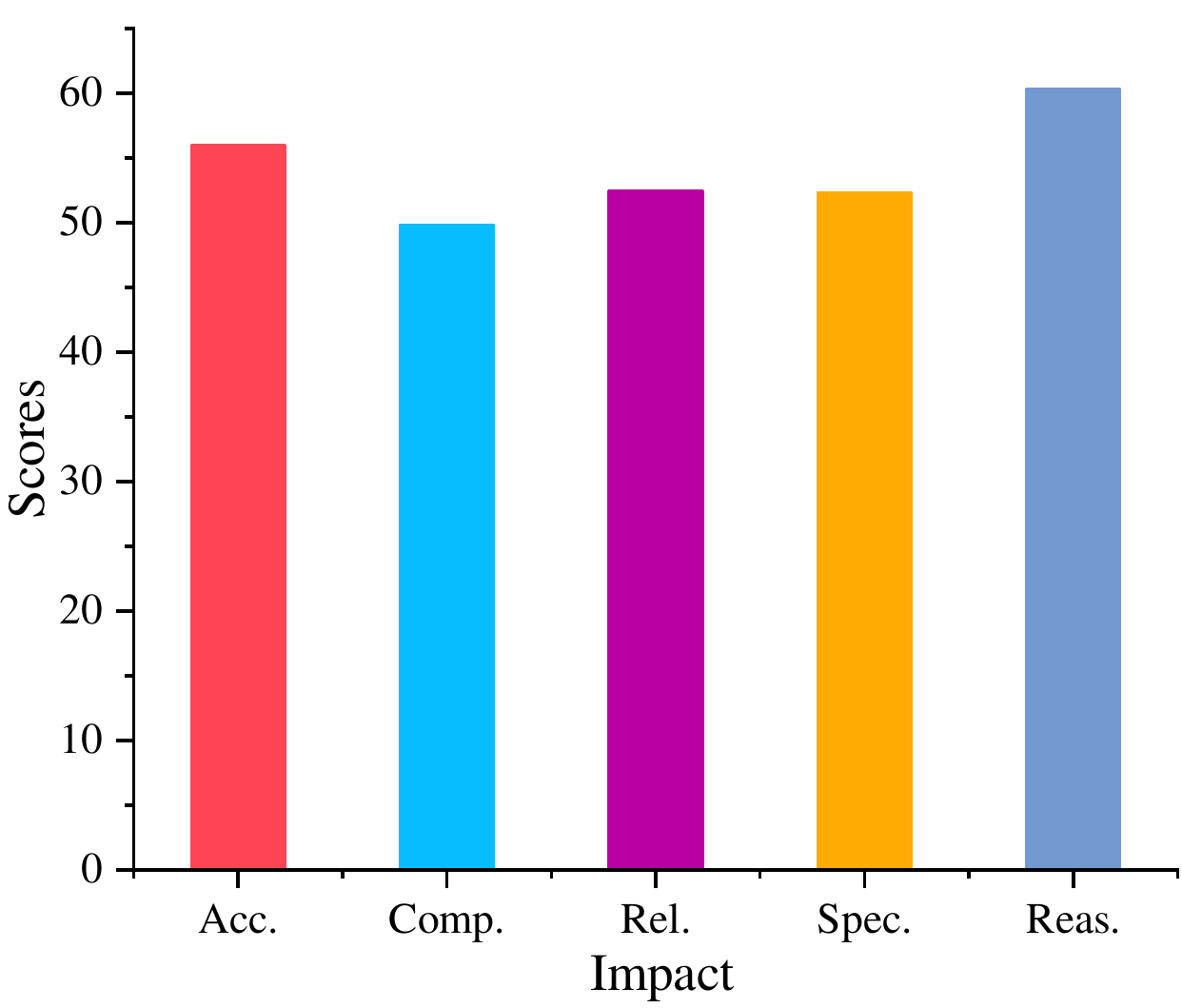}
\end{minipage}
\hfill
\begin{minipage}[t]{0.28\linewidth}
   \includegraphics[width = 1\linewidth]{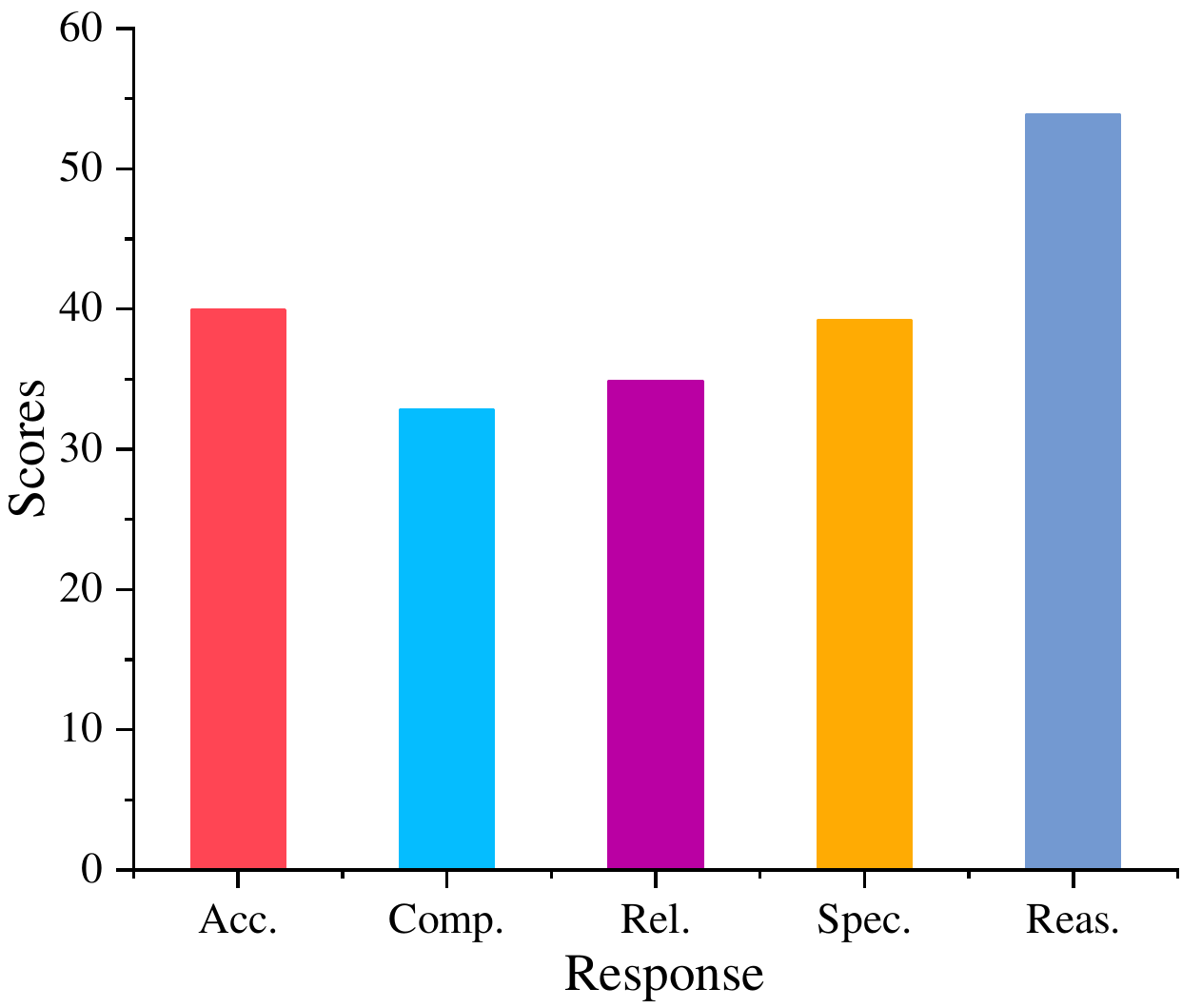}
\end{minipage}
\hfill
\begin{minipage}[t]{0.28\linewidth}
   \includegraphics[width = 1\linewidth]{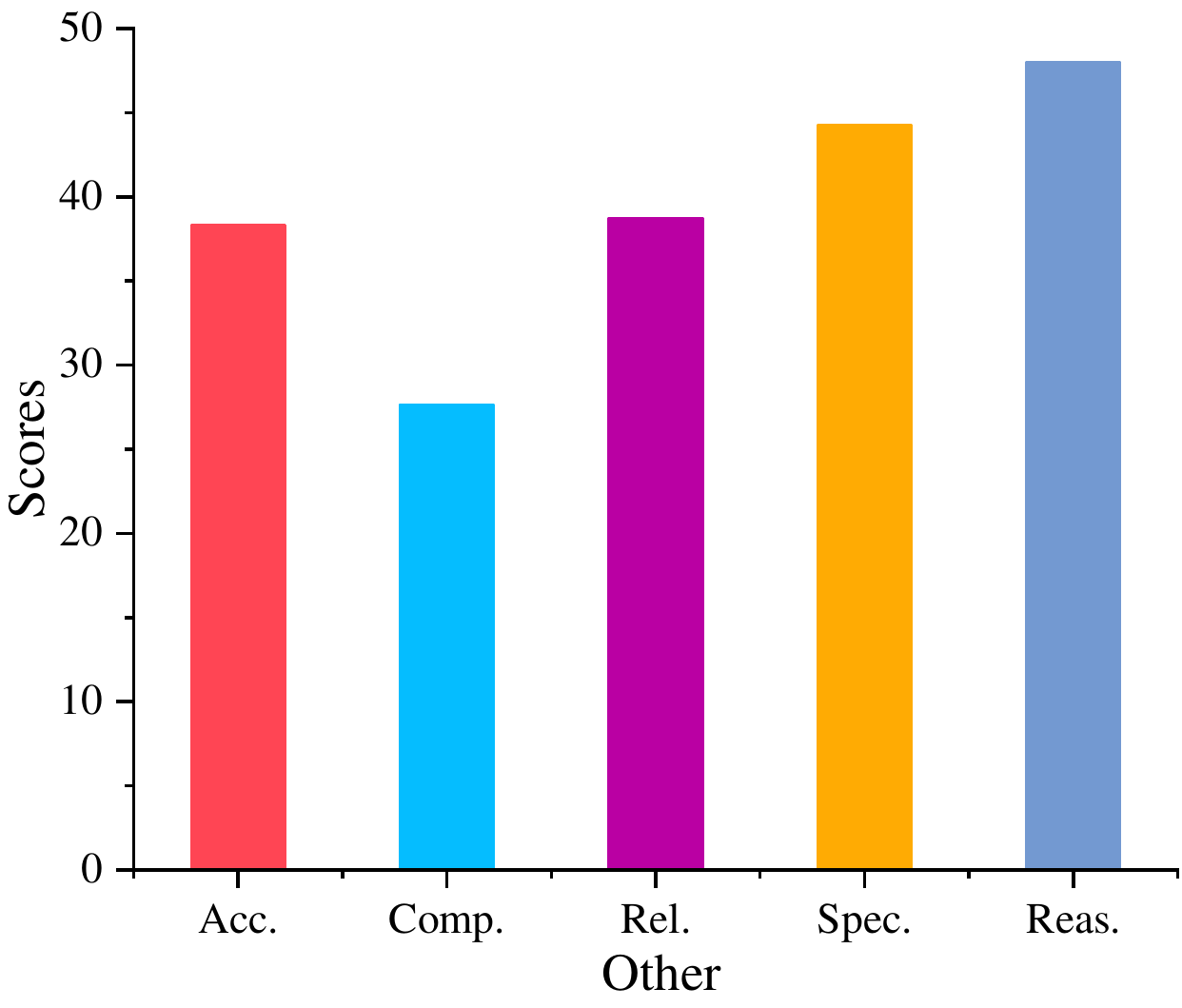}
\end{minipage}
\end{tabular}
\caption{GLM-4 Performance of individual dimension on Event Impact, Event Response, and Other.}
\label{fig:glm_indi_dim_2}
\end{figure*}

This section aims to provide a detailed analysis of the specific scores across different question types for each evaluation dimension. Figures \ref{fig:chat_indi_dem_1} and \ref{fig:chat_indi_dem_2} display the experimental results for GPT-3.5, while Figures \ref{fig:glm_indi_dim_1} and \ref{fig:glm_indi_dim_2} present the results for GLM-4. Overall, the experimental outcomes exhibit similar trends across both LLMs. Completeness scores are the lowest, indicating that making comprehensive future predictions is highly challenging. Reasonableness scores are relatively higher, suggesting that the predictions generated by the large models are logically consistent.

\begin{figure*}
    \centering
    \includegraphics[width=0.8\linewidth]{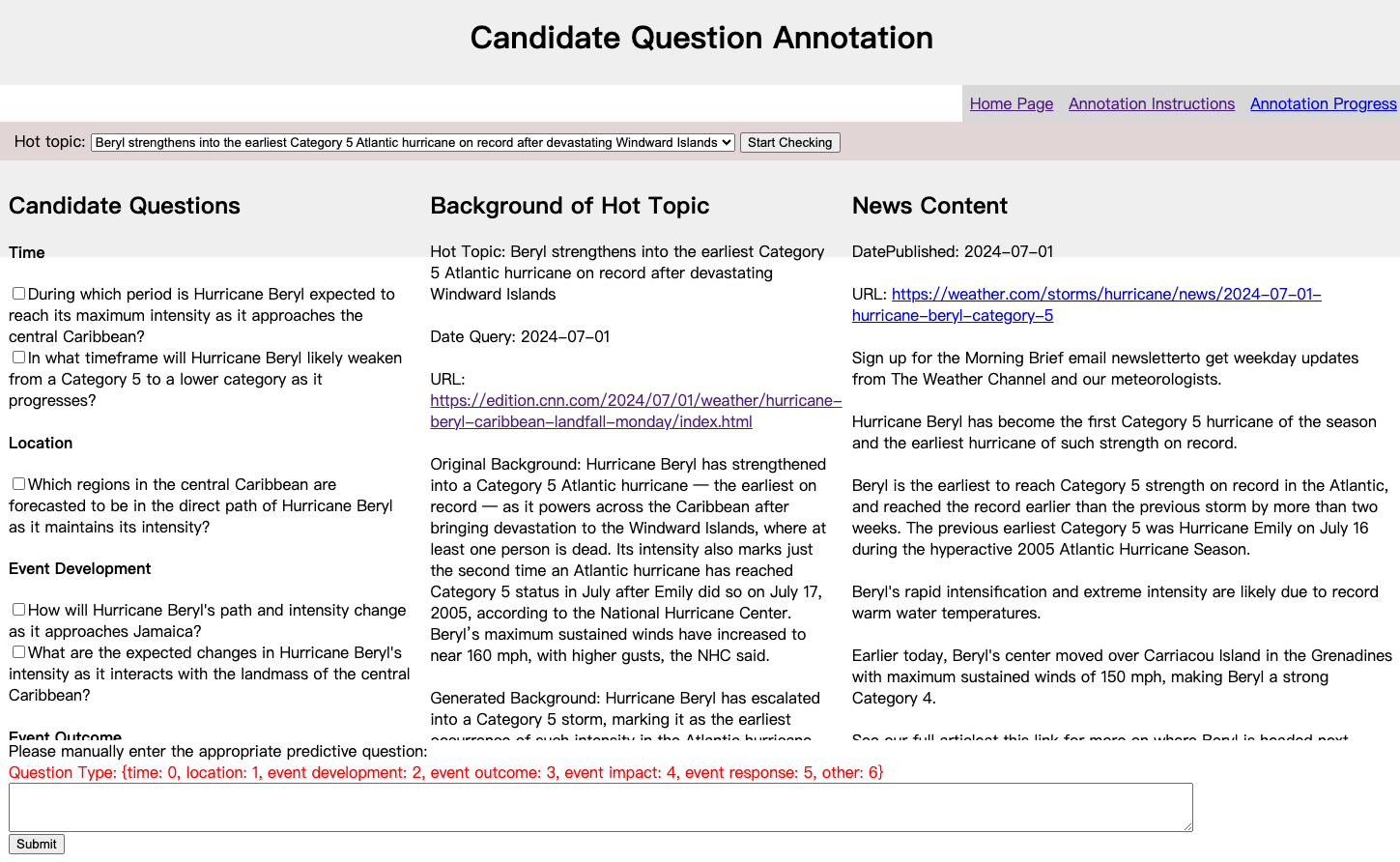}
    \caption{Question annotation interface.}
    \label{fig:question_anno}
\end{figure*}

\begin{figure*}
    \centering
    \includegraphics[width=0.8\linewidth]{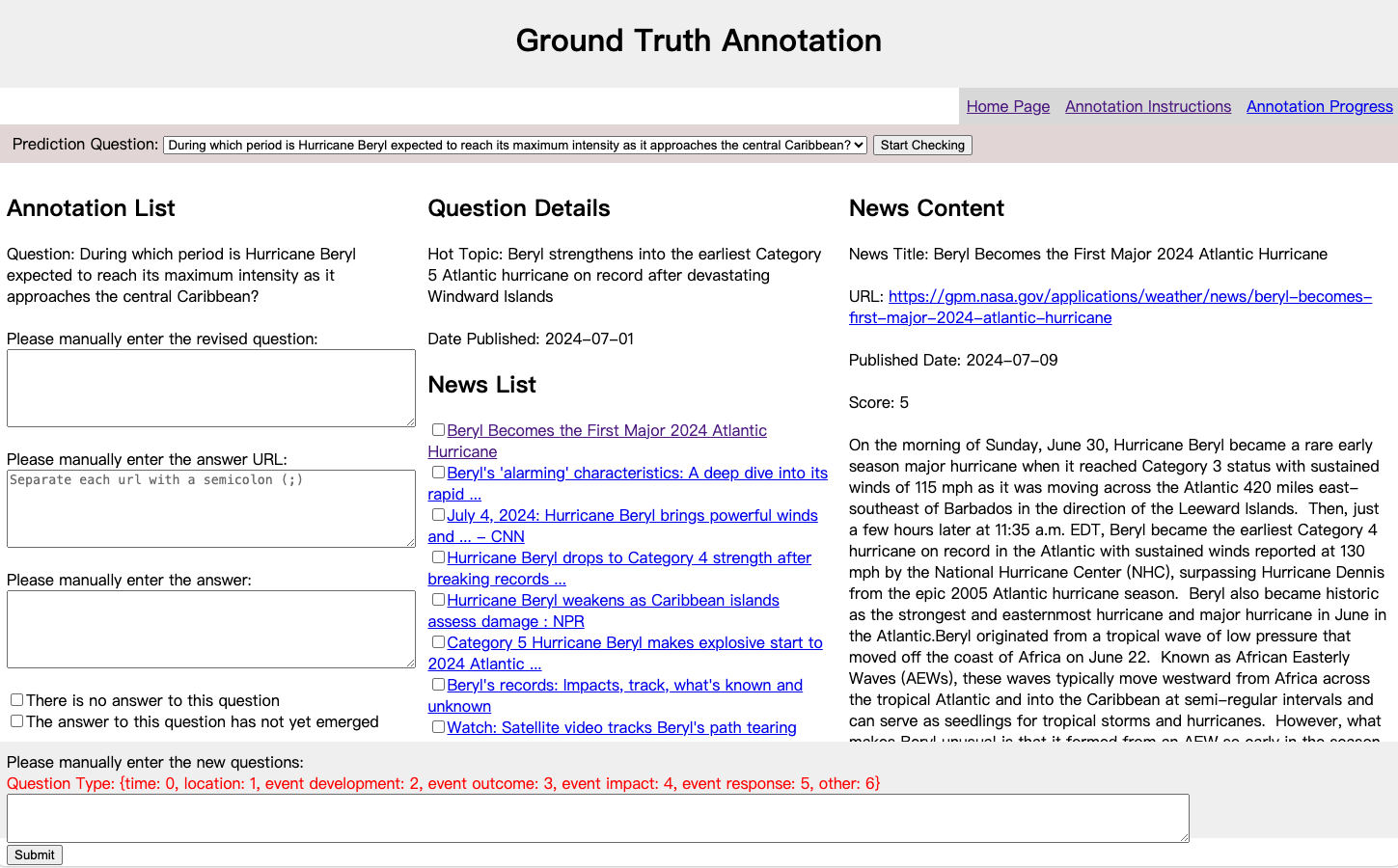}
    \caption{Ground truth annotation interface.}
    \label{fig:gt_anno}
\end{figure*}

\section{Details of Dataset Construction}
\label{appendx_data_construction}

We design the following six principles to better assist LLMs and human annotators in constructing the data.

(1) \textbf{Real-time Principle}. Events must be currently occurring. Data related to events not happening in real-time should be discarded, with potential scenarios including: 
(a) An event that occurred years ago has become a hot topic, such as ``\textit{A female employee was fired for using an umbrella at work to avoid exposure, and the court ruled the company's termination legal}''. 
(b) An event that happened some time ago and has been a hot topic for a while, such as ``\textit{How should one evaluate the role of a full-time postdoctoral fellow at Sichuan University}''. 

(2) \textbf{Answerability Principle}. For a predictive question, it must be ascertainable and answerable to warrant annotation. Unanswerable questions should be discarded. For example, ``\textit{How might the performance of the Chinese team in the next 15 days affect its status in international football?}'' On one hand, a few matches alone are insufficient to determine impacts on international status. On the other, the outcomes of such questions may not become apparent within 15 days and should therefore be discarded.

(3) \textbf{Specificity Principle}. Vague questions and those with broad, indeterminate answers should be discarded. For example, ``\textit{What impact might the STSS epidemic have on Japan in the next 15 days?}'' This question is unclear as the impacts could span multiple aspects, including economic and political, and should therefore be discarded.

(4) \textbf{Continuity Principle}. An event must still be unfolding and not concluded to justify its annotation. Events that have already ended should be discarded.

(5) \textbf{Short-Term Principle}. The current task of future event prediction primarily predicts events that may occur within the next 15 days. Therefore, it is necessary to analyze whether a predictive question can yield results within this prediction window. For example, ``\textit{What new laws might be proposed in response to construction safety incidents?}'' Legislative proposals typically do not yield results within 15 days and should be discarded.

(6) \textbf{Truthfulness Principle}. Events that are annotated must be real and currently occurring. People may pose discussions about events that have not actually happened. For example, ``\textit{Is there a future in opening all-female nursing homes for older single women?}'' or ``\textit{Do you remember what you did on the night after the college entrance exam ended?}''.

Furthermore, we have developed an annotation system. The question annotation interface is depicted in Figure \ref{fig:question_anno}, and the ground truth annotation interface is shown in Figure \ref{fig:gt_anno}.

\end{document}